\documentclass[10pt]{article} 
\PassOptionsToPackage{table,dvipsnames,svgnames}{xcolor}
\usepackage[preprint]{ml_format}

\usepackage[utf8]{inputenc} 
\usepackage[T1]{fontenc}    
\usepackage{hyperref}       
\usepackage{url}            
\usepackage{booktabs}       
\usepackage{amsfonts}       
\usepackage{nicefrac}       
\usepackage{microtype}      

\usepackage{amsmath}
\usepackage{amssymb}
\usepackage{mathtools}
\usepackage{amsthm}
\usepackage{basecommon} 

\colorlet{blueHighlightColor}{blue!05} 
\colorlet{wrongPredictionColor}{red!05} 
\colorlet{correctPredictionColor}{green!05} 
\definecolor{lightestgray}{gray}{0.95}

\usepackage{caption}
\usepackage{listings}
\newenvironment{ListingWithCaption}[1][]{\captionsetup{justification=raggedright, singlelinecheck=false}\captionsetup{type=lstlisting}\begin{minipage}{\linewidth}}{\end{minipage}\par}

\theoremstyle{plain}
\newtheorem{theorem}{Theorem}[section]

\theoremstyle{definition}
\newtheorem{definition}[theorem]{Definition}
\newtheorem{assumption}[theorem]{Assumption}
\theoremstyle{remark}
\newtheorem{remark}[theorem]{Remark}

\usepackage{algorithm}
\usepackage[noend]{algpseudocode}  
\algrenewcommand\alglinenumber[1]{\small #1:}

\title{Similarity-Distance-Magnitude Universal Verification}

\author{\name Allen Schmaltz \email allen@re.express \\
      \addr Reexpress AI}

\begin{document}

\maketitle

\begin{abstract}
We address the neural network robustness problem by adding \textsc{Similarity} (i.e., correctly predicted depth-matches into training)-awareness and \textsc{Distance}-to-training-distribution-awareness to the existing output \textsc{Magnitude} (i.e., decision-boundary)-awareness of the softmax function. The resulting \underline{\textsc{sdm} activation function} provides strong signals of the relative epistemic (reducible) predictive uncertainty. We use this novel behavior to further address the complementary HCI problem of mapping the output to human-interpretable summary statistics over relevant partitions of a held-out calibration set. Estimates of prediction-conditional uncertainty are obtained via a parsimonious learned transform over the class-conditional empirical CDFs of the output of a final-layer \textsc{sdm} activation function. For decision-making and as an intrinsic model check, estimates of class-conditional accuracy are obtained by further partitioning the high-probability regions of this calibrated output into class-conditional, region-specific CDFs. The uncertainty estimates from \underline{\textsc{sdm} calibration} are remarkably robust to test-time distribution shifts and out-of-distribution inputs; incorporate awareness of the effective sample size; provide estimates of uncertainty from the learning and data splitting processes; and are well-suited for selective classification and  conditional branching for additional test-time compute based on the predictive uncertainty, as for selective LLM generation, routing, and composition over multiple models and retrieval. Finally, we construct \underline{\textsc{sdm} networks}, LLMs with uncertainty-aware  verification and interpretability-by-exemplar as intrinsic properties. We provide open-source software implementing these results.\footnote{\url{https://github.com/ReexpressAI/sdm}}
\end{abstract}

\section{Introduction}

Large language models (LLMs) pose a challenge for interpretable and reliable deployment given the non-identifiability of their parameters \citep[inter alia]{hwang-and-ding-1997}\footnote{Informally, this means that two or more distinct sets of values for the parameters can result in identical output distributions. As a consequence, interpreting the parameters of such models is typically much more complicated than with a simple linear regression model, for example.}, which can number in the billions or more. Instead of directly interpreting parameters, instance-based, metric-learner approximations and hard-attention mechanisms can be constructed with task-specific inductive biases for effective semi-supervised learning (i.e., feature detection) and introspection against the training set \citep{schmaltz-2021-insights}, which can be useful for auditing predictions as a form of interpretability by example, or \textit{exemplar}, over the representation space of the model. However, for real-world deployments, robust approaches for predictive uncertainty---and relatedly, for verifying the modeling process---are also needed, both for human decision-making and for constructing sequentially dependent LLM pipelines.

Known theoretical results limit the statistical quantities that can be derived over LLMs. Statistical assurances in the distribution-free setting are limited to approximately conditional quantities \citep[inter alia]{Valiant-1984-PAC,LeiAndWasserman-2014-PredictionBands,BarberEtAl-2020-LimitsOfDistributionFree}. Further, even typical approximately conditional quantities can be difficult to obtain in practice, since the minimal assumption of exchangeability with a known held-out data set is itself often violated with co-variate and label shifts, which can be difficult to foresee with existing methods. Epistemologically, the prevalence of hallucinations and highly-confident wrong answers with widely deployed LLMs suggests a technical impasse in effectively modeling the predictive uncertainty, despite significant work from Bayesian, Frequentist, and empirically motivated perspectives \citep[inter alia]{GalAndZoubin-2016-MCDropout,Angelopoulos-2021-RAPS,GuoEtAl-2017-TempScaling,Lakshminarayanan-2017-DeepEnsembles,Ovadia-EtAl-2019-EvaluatingUncertainty}. A foundational piece is evidently missing from the picture.

Given these intrinsic challenges, we approach the problem of uncertainty quantification over LLMs from a new angle and ask: \textit{Can we leverage the metric learning and dense matching capabilities of neural networks over high-dimensional inputs to at least aim to maximize, with minimal distributional assumptions, the separation of aleatoric (irreducible) uncertainty and epistemic (reducible) uncertainty, decomposing the sources of the latter in a manner that is interpretable and actionable?}

We answer this question in the affirmative with a conceptually parsimonious, LLM-driven partitioning of the data to decompose sources of epistemic uncertainty: Correctly predicted depth-matches into the training set ($\Similarity$), the $\Distance$ to the training set, and the distance to the decision-boundary ($\Magnitude$). We use these signals to construct a new activation function, the $\sdm$ activation, which replaces a foundational building block of contemporary AI, the $\softmax$ operation. A series of distributional transforms over an $\sdm$ activation then enable us to directly target a quantity of interest, \textit{index-conditional calibration}, well-suited for selective classification \citep[inter alia]{Chow-1957-EarlyPredictWithRejectSystem,GeifmanAndEl-Yaniv-2017-NN-SelectiveClassification}, which reflects the typical need for uncertainty quantification with LLMs as part of multi-stage decision pipelines. Finally, with this new foundational behavior, we construct a new LLM architecture, the $\sdm$ network, with an intrinsic---and externally human interpretable---capability to verify its own instruction-following.  

In summary, in this work:
\begin{itemize}
\item We introduce the $\Similarity$-$\Distance$-$\Magnitude$ ($\sdm$) activation function, which encodes strong signals of epistemic uncertainty, to replace the $\softmax$ operation.
\item We provide a robust estimator of index-conditional uncertainty (Def.~\ref{defn:index-conditional-calibration}) via a final-layer $\sdm$ activation over existing models, unifying selective classification, calibration, and out-of-distribution detection for LLMs.
\item We propose the $\sdm$ network, a new LLM architecture and fine-tuning approach for which uncertainty-awareness and interpretability-by-exemplar are intrinsic properties. 
\item We empirically compare the uncertainty-awareness of the $\sdm$ estimator to existing classes of approaches, which we demonstrate do not reliably achieve our desired uncertainty quantity in the presence of---even modest---distribution shifts.
\item As a natural, held-out blind evaluation, we also demonstrate efficiently uncovering undetected annotation errors in the carefully curated \textsc{MMLU-Pro} benchmark dataset. This reflects the $\sdm$ estimator's capacity to separate aleatoric and epistemic uncertainty in high-probability regions.
\item More broadly, this work provides a new perspective on the behavior of neural networks, demonstrating that there are regions of the output distribution that are low variation and high probability that can be reliably detected. Existing modeling approaches marginalize over these regions, which can contribute to unexpected LLM behavior at test time.

\end{itemize}

\section{Motivation}

Given the ability of LLMs to recursively cross-encode data, user instructions, and outputs, if we had a reliable means of assessing the uncertainty over an LLM's predictions that was also human interpretable (i.e., a quantifiable and verifiable assurance in their instruction-following abilities), such an LLM could serve as a \textit{universal verifier} over existing models, which would in effect calibrate the predictive uncertainty of other models. For example, given an exogenous regression or multi-label model, one could simply cross-encode the data, exogenous model, and output as input to the LLM verifier and let the neural network generate the accuracy as to whether the exogenous model was correct or not. This process could be repeated, as needed, using such an LLM as a basis for building complex, compound AI systems, recursively cross-encoding the input and output, using the uncertainty over discrete predictions as the branching condition for additional test-time compute, tool calling, and human feedback --- and ultimately, reliable AI-assisted decision-making. In this work, we introduce the mechanisms for constructing such a verifier. 

\section{Preliminaries}

\subsection{Setting}
Both LLM next-token prediction and standard classification tasks (e.g., predicting the sentiment of a movie review) are formulated similarly as predictions over discrete classes.
We are given a training dataset, $\trainSplit=\{(\vx_n, y_n)\}_{n=1}^{N}$ of inputs, $\vx \in \gX$, paired with their corresponding ground-truth discrete labels, $y \in \gY = \{1, \ldots, C\}$, and a labeled calibration dataset, $\calibrationSplit$, drawn from the same distribution as $\trainSplit$. We are then given a new test instance, $\vx$, from an unlabeled test set, $\testSplit$, and seek to estimate the label with a prediction, $\hat{y}$, via the un-normalized log probabilities (``logits'', informally) of a final linear layer: $\vz  = \mW^T \vh + \vb$, where $\vh = \underlyingNetwork(\vx; \theta)$ is the final hidden state of a network parameterized by $\theta$. The network can be recurrent \citep{Hochreiter-and-Schmidhuber-1997-LSTM}, convolutional \citep{DauphinEtAl-2017-GCN}, or self-attention-based \citep{DevlinEtal-2019-BERT}, among others. The discrete prediction is taken as $\hat{y} = \argmax{\vz}$; however, for learning $\theta$, $\mW$, and $\vb$, and for human decision-making, we also seek an estimate of the predictive uncertainty, $p(y \given \vx)$, which is typically obtained by normalizing $\vz$ via the $\softmax$ operation described next. We will make a distinction between models, $\gM$ (defined by $\theta$, $\mW$, and $\vb$, and when applicable, the exemplar adaptor, described below), which produce the prediction, $\hat{y}$, and estimators, $\gE$, which provide an estimate of $p(y \given \vx)$, because different estimators can be used over the same model.

\subsection{Softmax and the Cross-Entropy loss}

The $\softmax$ has as its origins the work of L. Boltzmann in the 19th century \citep[see][]{SharpAndMatschinsky-2015-Boltzmann1877Translation}. It remains a central function in the natural and engineering sciences. It is ubiquitous in deep learning, playing an integral role as a router in self-attention mechanisms \citep{VaswaniEtAl-2017-AttentionIsAllYouNeed} and mixture-of-experts models \citep{ShazeerEtAl-2017-MOE}; forming the basis of the cross-entropy loss used for next-token training of LLMs; and serving as the final interface between a model and the end-user, converting the un-normalized model logits to human interpretable probability distributions, at least in principle:
\begin{align}\label{eq:softmax}
\softmax(\vz)_i = \frac{
e^{\tau \cdot z_i}
}{
\sum^C_{c=1}{e^{\tau \cdot z_c}}
}, 1 \le i \le C, \tau \ge 0
\end{align}
The above function induces a parameterization of the event probabilities of a categorical distribution:
\begin{align}\label{eq:softmaxCategorical}
\CategoricalDistribution(C=|\gY|, \softmax(\vz )) 
\end{align}
The inverse-temperature parameter, $\tau$, controls the sharpness of the distribution. As $\tau \rightarrow 0$, the output of $\softmax(\vz)$ converges to a uniform distribution where each class has probability $\frac{1}{C}$; as $\tau \rightarrow \infty$, the output converges to a distribution in which all of the mass is assigned to a single class. In deep learning, $\tau$ is treated as a learnable, \textit{global} hyper-parameter; \textit{instance-wise} variation in the distance to the decision-boundary is thus determined by the relative $\Magnitude$ of $z_{\hat{y}}$. This model is learned by minimizing the cross-entropy loss between $\vz$ and the index of the true labels over $\trainSplit$. The \textit{natural} logarithm of the loss is the counterpart to the base $e$ of the $\softmax$:
\begin{align}\label{eq:cross-entropy-loss}
\gL(\theta, \mW, \vb; \trainSplit) = -\frac{1}{N} \sum_{n}^{N} \log_{e}\left(
\frac{
e^{\tau \cdot z_{y_n}}
}{
\sum^C_{c=1}{e^{\tau \cdot z_c}}
}
\right)
\end{align}
\section{Methods}
\begin{figure}[t]
\centering
\includegraphics[width=0.75\textwidth]{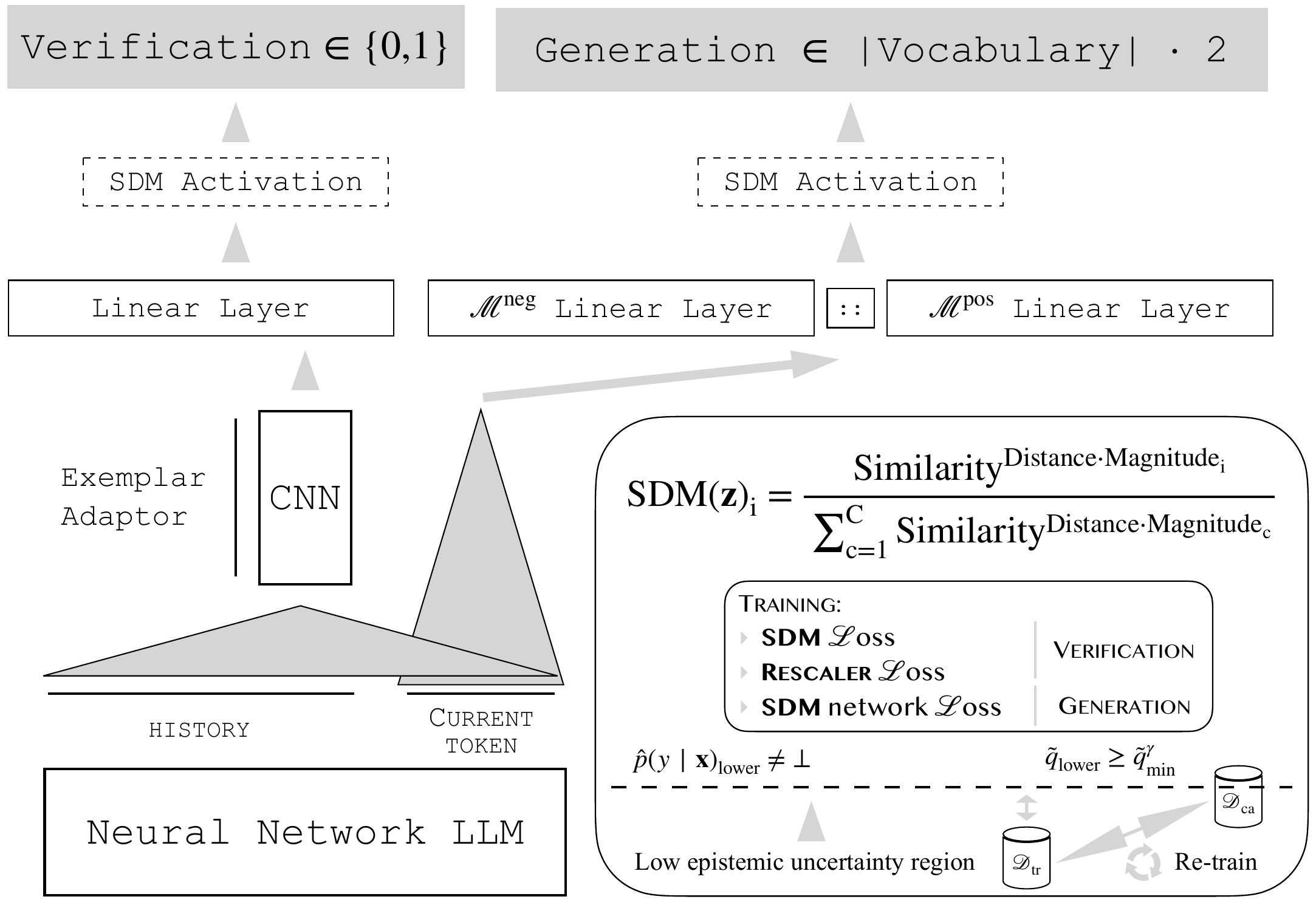}
\caption{$\sdm$ networks are uncertainty-aware via a robust estimator of index-conditional calibration (Def.~\ref{defn:index-conditional-calibration}), $\EstimatorPLower$, over output verification (i.e., binary classification of instruction-following); intrinsically introspectable via depth-matching into a training set ($\trainSplit$) and correspondence to comparable points in a held-out calibration set ($\calibrationSplit$) via $\hardqbin$, which is a stable mapping and summary of the epistemic uncertainty signals of $\Similarity$, $\Distance$, and $\Magnitude$; and updatable via a fine-tuning process to maximize the proportion of verifiable high-probability generations. Decoding proceeds by generating from the distribution of $\sdm(\vz_{\rm{neg}}, \vz_{\rm{pos}})$ up to a control token at the unit-of-analysis of the verification labels. Decoding then continues, or other branching actions are taken, based on $\EstimatorPLower$.}
\label{fig:sdm-network-diagram}
\end{figure}
In this work, we revisit Eq.~\ref{eq:softmax},~\ref{eq:softmaxCategorical}, and ~\ref{eq:cross-entropy-loss} given new observations on the statistical behavior of high-dimensional objects, empirically derived from large parameter neural networks. We will seek to decouple the sources of epistemic uncertainty via a new activation function that is conceptually: 
\begin{align}\label{eq:abstractSMD}
\sdm(\vz)_i = 
\frac{
{\Similarity}^{\Distance \cdot \Magnitude_i}
}{
\sum^C_{c=1}{{\Similarity}^{\Distance \cdot \Magnitude_c}}
}
\end{align}
with a corresponding negative log likelihood loss that takes into account the change of base (\S~\ref{sec:sdm-activation}). We will additionally introduce a transformation that rescales this value for an \textit{instance} with exogenous information \textit{across} $\calibrationSplit$, effectively calibrating \citep{Brier-1950-BrierCalibration,Dawid-1982-CalibratedBayesian} the model to produce reliable, interpretable probabilities (\S~\ref{sec:sdm-calibration}). Finally, we integrate this behavior into the LLM architecture and training, yielding an LLM with an intrinsic ability to verify its own instruction following (\S~\ref{sec:sdm-networks}), as illustrated in Figure~\ref{fig:sdm-network-diagram}. 

\subsection{From Model Approximations via Exemplar Adaptors to SDM Activation Functions}\label{sec:sdm-activation}

Exemplar adaptors, 1-D CNN adaptors (with a final linear layer) over the frozen hidden states of a network, induce distilled, compressed representations of an underlying network's representation space conditional on its predictions. This behavior can be used to faithfully approximate a model's predictions as a mapping against a training, or support, set. This can be achieved, for example, with instance-based, metric-learning estimators, such as weighted KNNs, where the weights are learned as a transform of the exemplar adaptor's distilled representations.\footnote{
Such instance-based, metric-learner approximations of neural networks differ from traditional KNN rules \citep[inter alia]{CoverAndHart-1967-KNNBayesError,DevroyeEtAl-1996-APT} in two critical respects: The neural network serves as a semi-supervised learner of the distances between the dense representations that identify the instances, and there is a model prediction (in addition to the ground-truth label) for each instance in the support set. The former enables effective partitioning despite the curse of high dimensions; the latter provides an additional indicator of reliability for each instance.
} Critically, when the approximations diverge from the predictions of the underlying model, the inputs tend to be from the subsets of the distribution over which the underlying model is itself unreliable \citep{schmaltz-2021-insights}. In other words, the approximations encode strong signals of the epistemic uncertainty. Rather than constructing explicit KNN approximations, which require a separate training step and additional parameters, we instead \textbf{q}uantize the closeness of a point to the training set with a discrete estimate. Further, we transform the distance to the closest match as a quantile estimate over the distribution of distances. These quantities, combined with the output $\Magnitude$, capture the key sources of epistemic uncertainty for an input \textit{instance} (cf. \S~\ref{sec:sdm-calibration}).
\subsubsection{Exemplar Adaptor}
We take as the CNN of our exemplar adaptor $g: (\vh, t(\vz)) \in \reals^D \mapsto \vh' \in \reals^M$, a 1-D CNN that takes as input $h$ (if available) of the underlying network and optionally, the concatenation of the output of $t(\vz)$, a transform of the underlying network's output.\footnote{
For black-box LLM API's in particular, we will not have direct access to $\vh$ and will instead construct a proxy of $\vh$ via a transform $t$ of the available output, which may (and typically will with current models) itself be the result of a $\softmax$ operation.
}
The CNN has $M$ filters, the filter applications of which produce $\vh'$, the distilled representation of the underlying network. A final linear layer, $\vz'  = \mW'^T \vh' + \vb', \vz' \in \reals^C$, then replaces the underlying network's linear layer, with the discrete prediction taken as $\hat{y} = \argmax{\vz'}$. This exemplar adaptor will then enable us to derive the key signals of epistemic uncertainty, $\Similarity$, $\Distance$, and $\Magnitude$ described next.

\subsubsection{$\Similarity$} 

We define the $\Similarity$ ($\q$) of an instance to the training set as the count of consecutive nearest matches in $\trainSplit$ that are correctly predicted \textit{and} match $\hat{y}$ of the test instance. Concretely, we first sort $\gD_{\rm{tr}}$ (for which we have both model predictions and ground-truth labels) based on the $L^2$ distance (2-norm) from $\vh'$, $\left[(\vx^{tr}_{(1)}, \hat{y}^{tr}_{(1)}, y^{tr}_{(1)}),\ldots, (\vx^{tr}_{(N)}, \hat{y}^{tr}_{(N)}, y^{tr}_{(N)})\right]$, such that $|| \vh' - \vh'^{tr}_{(1)} ||_2 \le \ldots \le || \vh' - \vh'^{tr}_{(N)} ||_2$, and then calculate $q \in \{0, \ldots, | \trainSplit |\}$ as:
\begin{equation}\label{equation:q} 
q = 
\sum^{| \trainSplit |}_{i=1} \mathbf{1}_{\hat{y} =  \hat{y}^{\rm{tr}}_{(i)}} \cdot \mathbf{1}_{\hat{y}^{\rm{tr}}_{ (i) } = y^{\rm{tr}}_{ (i) }} \cdot \mathbf{1}_{
i-1 = \sum^{i-1}_{j=1} \mathbf{1}_{\hat{y} =  \hat{y}^{\rm{tr}}_{(j)}} \cdot \mathbf{1}_{\hat{y}^{\rm{tr}}_{ (j) } = y^{\rm{tr}}_{ (j) }}
}
\end{equation}
where the rightmost indicator function, $\mathbf{1} \in \{ 0, 1\}$, ensures consecutive (depth-wise) matches. By definition, $\q$ cannot exceed the count of the most prevalent class label in $\trainSplit$, and since we assume an approximately equal number of points for each class, $\q \ll | \trainSplit |$ is typical. For the special case of calculating $q$ for $\vx \in \trainSplit$, which only occurs during learning, we exclude the self-match.

\subsubsection{$\Distance$}
The $L^2$ distance to the nearest match in $\trainSplit$ follows from above: $\dNearest = || \vh' - \vh'^{tr}_{(1)} ||_2$. However, it is difficult to work with $\dNearest$ directly since its scale can vary widely depending on the input to $g$ and the size of $M$. Instead, we define $\Distance$, $\dVal \in [0,1]$, in terms of the class-wise empirical CDFs of $\dNearest$ over $\calibrationSplit$, as the most conservative quantile relative to the distance to the nearest matches observed in the labeled, held-out set:
\begin{equation}\label{equation:d} 
\dVal = \min\left [ 1-{\rm{eCDF}}^{y_1}_{\rm{ca}}( \dNearest ) , \ldots , 1-{\rm{eCDF}}^{y_C}_{\rm{ca}}( \dNearest ) \right ]
\end{equation}
The empirical CDFs are determined by the labeled points in $\calibrationSplit$ for which $\q>0$, where, as indicated by the superscripts, the stratification of points is by the true labels, $y$. For example, ${\rm{eCDF}}^{y_1}_{\rm{ca}}( \dNearest )$ is the empirical CDF of $\dNearest$ values in $\calibrationSplit$ for which $y=1$, a notation convention we will use throughout. (Points with $\q=0$ are effectively out-of-distribution points and treated as such in downstream decision-making, so they are excluded to avoid biasing the estimates.) At test time, we do not see $y$; instead, the minimum is calculated over the quantiles of each of the class-conditional eCDFs, regardless of $\hat{y}$. As with $q$, for the special case of calculating $\dVal$ for $\vx \in \trainSplit$, we replace ${\rm{eCDF}}^{y_c}_{\rm{ca}}$ with the analogous ${\rm{eCDF}}^{y_c}_{\rm{tr}}$, the class-wise empirical CDFs of $\dNearest$ over $\trainSplit$ excluding self-matches.
\subsubsection{$\Magnitude$}
We take as the $\Magnitude$, or distance to the decision boundary, $z'_{\hat{y}}$, as in the standard $\softmax$ case but via $\vz'$ from the linear layer of the exemplar adaptor. 

\subsubsection{SDM Activation: Formulation}
We use the above quantities to define the $\sdm$ activation function:
\begin{align}\label{eq:sdmActivation}
\sdm(\vz')_i = \frac{
(2+q)^{\dVal \cdot z'_i}
}{
\sum^C_{c=1}{(2+q)^{\dVal \cdot z'_c}}
}, 1 \le i \le C
\end{align}
The output distribution becomes sharper with higher values of $\q$, $\dVal$, and $z'$. Also note that when $\dNearest$ exceeds the largest distance observed in the labeled data, $\dVal=0$ and the output distribution is uniform, reflecting a maximally high (i.e., out-of-distribution) epistemic uncertainty estimate. The standard $\softmax$ with $\tau=1$ is recovered by setting $q = e-2, d=1$. As with the $\softmax$ operation, $\argmax \sdm(\vz') = \argmax \vz'$.
 
\subsubsection{SDM Activation: Loss and Training}

A loss analogous to Eq.~\ref{eq:cross-entropy-loss} then follows with the applicable change of base. We use this loss to train the weights of the exemplar adaptor, which includes the parameters of the linear layer ($\mW'$ and $\vb'$), as well as the convolution weights and biases, which we collectively represent with $\mG$. The weights of the underlying $\underlyingNetwork$ remain fixed. (We return to training $\theta$, $\mW$, and $\vb$ of an underlying LLM in \S~\ref{sec:sdm-networks}.)
\begin{align}\label{eq:sdm-loss}
\gL(\mG, \mW', \vb'; \trainSplit) = -\frac{1}{N} \sum_{n}^{N} \log_{(2+q)}\left(
\frac{
(2+q)^{\dVal \cdot z'_{y_n}}
}{
\sum^C_{c=1}{(2+q)^{\dVal \cdot z'_c}}
}
\right)
\end{align}

Pseudo-code for training the $\sdm$ activation layer and $\sdm$ estimator (described in \S~\ref{sec:sdm-calibration}, next) appears in Alg.~\ref{alg:sdm-estimator-training}. The first epoch is initialized with a standard $\softmax$ (i.e., setting $q = e-2, d=1$). Training then proceeds by re-estimating $q$ and $d$ for each $\vx \in \trainSplit$ after each epoch. We take as the stopping criteria for one learning round as the epoch with the highest average balanced (across classes) median $q$ values over $\calibrationSplit$. We choose the final model $\gM_{*} \in \sM$ over $J$ iterations of random shuffles and splits of $\trainSplit$ and $\calibrationSplit$ and parameter initializations as that with the globally highest average balanced (across classes) median $q$ values over $\calibrationSplit$. For learning, we assume $\trainSplit$ and $\calibrationSplit$ are balanced across all class labels, $c \in \gY$.

\subsection{From SDM Activation Functions to SDM Calibration}\label{sec:sdm-calibration}

Given a fixed underlying $\underlyingNetwork$, the \textsc{sdm} activation function in Eq.~\ref{eq:sdmActivation} encodes strong signals of the epistemic uncertainty of a single instance for a single model $\gM_{*} \in \sM$, but a priori, it is not sufficient alone for calibration without additional exogenous information, since it does not explicitly take into account the epistemic uncertainty from the splitting of $\trainSplit$ and $\calibrationSplit$; the stochasticity of parameter initialization; and the stochasticity of the learning process, more generally. Relatedly, to enable the interpretability of the calibration process (e.g., to perform model checks), we need a stable mapping of test points to the relevant partitions of $\calibrationSplit$. 

In service of achieving these additional properties, we first need to specify a definition of calibration, of which there are conflicting quantities, definitions, and evaluation metrics \citep{VaicenaviciusEtAl-2019-EvaluatingCalibration,KullEtAl-2019-BeyondTempScaling,GuptaAndRamdas-2022-ToplabelCalibration}. Fortunately, in real-world settings with LLMs, we are primarily concerned with reliably detecting high-probability regions, which significantly simplifies the evaluations and removes much of the ambiguity in the definitions. To motivate our definition, we first consider two under-specified definitions of calibration, in which the true long-run frequencies of the ground-truth labels match the probability estimates from the estimator, $\gE$, stratified by the predicted class, $\hat{y}$, and the true class, $y$, respectively, given some un-specified binning of the real-valued probabilities:

\begin{definition}\label{defn:prediction-conditional-calibration}
An estimator, $\gE$, of $p(y \given \vx)$ is \textit{prediction-conditional calibrated}, if $\forall~\alpha' \in [0,1]$:
$p(y = \hat{y} \given \hat{y}, \gE(\vx) = \alpha') = \alpha'$.
\end{definition}

\begin{definition}\label{defn:class-conditional-calibration}
An estimator, $\gE$, of $p(y \given \vx)$ is \textit{class-conditional calibrated}, if $\forall~\alpha' \in [0,1]$:
$p(y = \hat{y} \given y, \gE(\vx) = \alpha') = \alpha'$.
\end{definition}

Assuming no distribution shifts, and setting aside conditioning on additional attributes and the method of binning, the source of the under-specification, Def.~\ref{defn:class-conditional-calibration} is a generally more informative quantity, but cannot be meaningfully estimated across all points since the true label, $y$, is not available at test time. Thus, calibration becomes a tension between the quantities desired and the regions---and the size (sharpness) of those regions---that can be partitioned. Most works are premised on a variation of Def.~\ref{defn:prediction-conditional-calibration}; an alternative compromise is taken by frequentist conformal estimators by changing the quantity to coverage over a discrete prediction set. We will instead seek the following quantity, which aligns with the quantity needed for selective classification for conditional branching of LLM compute and final human decision-making dependent on the presence of high-probability predictions:

\begin{definition}\label{defn:index-conditional-calibration}
An estimator, $\gE$, of $p(y \given \vx)$ is \textit{index-conditional calibrated} at $\alpha' \in (\frac{1}{C}, 1]$ if:
$p(y = \hat{y} \given \hat{y}, \gE(\vx) \ge \alpha') \ge \alpha'$ $\wedge$ $p(y = \hat{y} \given y, \gE(\vx) \ge \alpha') \ge \alpha'$.
\end{definition}

To evaluate this quantity, we only consider the points for which the estimator assigns a high-probability of at least $\alpha'$, which is typically near 1, such as $1-\alpha=\alpha'=0.95$ in our experiments. We refer to this set of points as the \textit{admitted}, or \textit{non-rejected}, set. Then, given ground-truth values for $\testSplit$, we assess whether the conditional accuracies of the admitted set are at least $\alpha'$ when stratifying by the predicted labels, $\hat{y}$, and the true labels, $y$. Unlike evaluating Def.~\ref{defn:prediction-conditional-calibration}, there is thus no ambiguity with regard to the choice of binning the probabilities.

\begin{algorithm}
    \caption{$\sdm$ Activation Layer and $\sdm$ Estimator Training}
    \label{alg:sdm-estimator-training}
    \small
    \begin{algorithmic}[1] 
        \Require $\trainSplit$, $\calibrationSplit$, $\alpha'$, $\underlyingNetwork$, $\text{max epochs}$, $\text{rescaler max epochs}$, $\text{rescaler stopping condition}$ %
        \State \textbf{Assumption:} $\trainSplit$, $\calibrationSplit$ are balanced across all class labels, $c \in \gY$
        \Procedure{sdm-iterative-train}{$\trainSplit$, $\calibrationSplit$, $\alpha'$, $\underlyingNetwork$, $\text{max epochs}$}
        \State $\gM_{*} \gets \emptyset$ \Comment{Globally best model}
        \State $\trainSplit_{*} \gets \emptyset$, $\calibrationSplit_{*} \gets \emptyset$  \Comment{Data splits of best model}
        \State $\gE \gets \emptyset$  \Comment{$\sdm$ estimator (i.e., $\EstimatorPLower$)}
        \State ${\rm{metric}}_{*} \gets 0$ \Comment{Determines final best model}
        \State ${\rm{stats}} \gets \{~\}$ \Comment{Summary statistics to calculate $\CauchyMinValidQBin, \CauchyMagnitudeIteratedLowerEstimate$ (\S~\ref{sec:iterative-uncertainty})}
        \For{$j \in 1, \ldots, J$} \Comment{The learning process is repeated $J$ times}
		\State $\gM_{j*} = \emptyset$ \Comment{Best model for a single learning round}
		\State $\rm{metric}_j \gets 0$
		\State $\trainSplit$, $\calibrationSplit \gets$ Random shuffle and even split of $\trainSplit$ and $\calibrationSplit$
		\State $\gM_j \gets $ Random initialization of $\mG_j, \mW'_j, \vb'_j$
		\State $\q \gets e-2, d \gets 1$ \Comment{Standard $\softmax$ for first epoch}
		 \For{$e \in 1, \ldots, \text{max epochs}$}
		 	\State Minimize $\gL(\mG, \mW', \vb'; \trainSplit)$ \Comment{Eq.~\ref{eq:sdm-loss}}
			\State Update $q,d$ for each $\vx \in \trainSplit$
			\State $\rm{metric} \gets$ mean balanced (across $c \in \gY$) median $\q$ over $\calibrationSplit$
          	 	\If{$\rm{metric} \ge \rm{metric}_j$}
           			\State $\rm{metric}_j \gets \rm{metric}$
				\State $\gM_{j*} \gets \gM_j$
           		\EndIf
		        \If{$\rm{metric}_j \ge {\rm{metric}}_{*}$}
           			\State ${\rm{metric}}_{*} \gets \rm{metric}_j$
				\State $\gM_{*} \gets \gM_{j*}$
				\State $\trainSplit_{*}, \calibrationSplit_{*} \gets \trainSplit, \calibrationSplit$ \Comment{Data splits for calculating $q$ and $d$ at test time}
           		\EndIf
		 \EndFor
		 \State $\gM_{j*} \gets$ update with $\mW^{''}$ from \Call{train-rescaler}{$\cdot$} \Comment{ Alg.~\ref{alg:train-rescaler}}
		 \State ${\rm{stats}} \gets$ update with \Call{find-min-rescaled-q}{$\cdot$} \Comment{ Alg.~\ref{alg:estimate-soft-q-bin}}
        \EndFor
        \State $\gE \gets $ Constructed from globally best model $\gM_{*}$ (and associated values, e.g., $\minValidQBin_*$) and ${\rm{stats}}$
        \State \textbf{return} $\gM_{*}, \trainSplit_{*}, \calibrationSplit_{*}, \gE$
        \EndProcedure
        \Ensure $\gM_{*}, \trainSplit_{*}, \calibrationSplit_{*}, \gE$ 
    \end{algorithmic}
\end{algorithm}

The estimator that rejects all points is index-conditional calibrated. Given two estimators that are index-conditional calibrated, we prefer that which rejects fewer points, ceteris paribus. In other words, we seek estimators that meet our reliability condition and are informative (i.e., maximize the number of points that are properly admitted), but when the estimator is uncertain, we prefer rejection over unexpectedly falling under the desired $\alpha'$ probability threshold.

The key compromise is that we will not be able to reliably calculate a probability for all points; however, for LLM tasks, there is typically not an actionable notion of partial acceptability for final decision-making, so it is a reasonable compromise. Either the complex LLM output is verified as correct, or some separate, remedial action must be taken, such as dividing the task into simpler tasks, reformatting and re-cross-encoding, and/or retrieving information exogenous to the model, where again for each of these sub-tasks, we seek index-conditional calibrated estimators at the level of the available labels, where the stopping condition is eventually deferment to human adjudication.

Despite the aforementioned compromise, and although evaluation is unambiguous, it may still seem mysterious that the second condition of Def.~\ref{defn:index-conditional-calibration} can be meaningfully estimated. To do so, we will need to perform a series of transforms over the already strong uncertainty signals from the $\sdm$ activation function and re-visit the behavior of partitioning empirical CDFs, to which we turn next.

\subsubsection{Rescaling SDM Activation Output to Account for Effective Sample Sizes}\label{sec:rescale-for-effective-sample-size}

A disadvantage of using $\sdm(\vz')$ directly as an estimator is that it only has an indirect, relative notion of the effective sample size of $\calibrationSplit$. Intuitively, the confidence in a prediction should be commensurate with the number of comparable points in $\trainSplit$ and $\calibrationSplit$, which the $\sdm$ activation captures via $\Similarity$, $\Distance$, and $\Magnitude$. For example, an out-of-distribution point will tend to have $d=0$ and low values of $\q$, reflecting a small effective sample size in the observed data. However, to further improve the robustness of the estimate, we can explicitly incorporate an additional, direct notion of the effective sample size via distributional statistics over $\calibrationSplit$.

First, we calculate class-conditional empirical CDFs over $\calibrationSplit$ of the output of $\sdm(\vz')$. For a given point, this will create a vector, $\vv \in \reals^C$, of the quantiles:
\begin{equation}\label{equation:sdm-output-cdfs} 
\vv = \left [ {\rm{eCDF}}^{y_1}_{\rm{ca}}( \sdm(\vz')_1 ) , \ldots , {\rm{eCDF}}^{y_C}_{\rm{ca}}( \sdm(\vz')_C ) \right ]
\end{equation}
Next, we rescale $\q$ to take into account these distributional statistics. The resulting value will be the basis for our stable mapping between new, unseen test points and $\calibrationSplit$:
\begin{equation}\label{equation:softqbin} 
\softqbin = log_{e} \left ( (2+q)^{\vv_{\hat{y}}} \right )
\end{equation}

We seek a normalized distribution both to present to users and to enable the subsequent transform described in \S~\ref{sec:region-specific-ecdfs}. Toward this end, we rescale with a linear layer, without a bias, the training of which we detail in \S~\ref{sec:training-the-rescaler}: $\vv'  = \mW^{''T} \vv, \vv' \in \reals^C$. This is normalized using $2+\softqbin$ as the base, $\vo \in \reals^C$:
\begin{align}\label{equation:rescaled-prediction-conditional-estimate} 
o_i = \frac{
(2+\softqbin)^{~v'_i}
}{
\sum^C_{c=1}{(2+\softqbin)^{~v'_c}}
}, 1 \le i \le C
\end{align}
Unlike the output of an $\sdm$ activation, $\argmax \vo$ is not necessarily (but typically will be) equivalent to $\hat{y} = \argmax \vz'$. When they are not equivalent, our convention is to set $\softqbin = 0$ for the point, which will in effect treat the point as out-of-distribution in downstream analyses.

\paragraph{Effective Sample Sizes via the DKW Inequality.}
Eq.~\ref{equation:rescaled-prediction-conditional-estimate} is premised on the assumption that the empirical CDFs in Eq.~\ref{equation:sdm-output-cdfs} reflect the true, underlying conditional distributions, which are unspecified.\footnote{That is also true for the eCDFs in Eq~\ref{equation:d}, but we make the reasonable assumption that the higher-level transforms starting with Eq.~\ref{equation:sdm-output-cdfs} effectively account for the uncertainty in the distance eCDFs.} That would seem to be a relatively strong assumption as the final estimate, particularly for small sample sizes, even if empirically effective over existing datasets, and is the entry point for incorporating an explicit notion of the effective sample size in our estimates.

We make the following conservative assumption, parameterizing the prior belief that data points with a looser connection to $\trainSplit$ reflect smaller effective sample sizes, while also explicitly accounting for the count of observed points in $\calibrationSplit$:
\begin{assumption}
We assume the effective sample size is increasing in $\softqbin$, class-wise over $\calibrationSplit$.
\end{assumption}
For each $\vx \in \testSplit$, using $\softqbin$, we calculate the vector of effective sample sizes across classes, $\mathbf{\hat{n}}$, relative to $\calibrationSplit$ as: 
\begin{align}\label{equation:effective-sample-size} 
\mathbf{\hat{n}} = \left [ | \calibrationSplit |^{y_1} \cdot {\rm{eCDF}}^{y_1}_{\rm{ca}}( \softqbin ) , \ldots , | \calibrationSplit |^{y_C} \cdot {\rm{eCDF}}^{y_C}_{\rm{ca}}( \softqbin ) \right ]
\end{align}
where $ | \calibrationSplit |^{y_c}$ is the count of calibration set points with true label $y = c$.

With these sample size estimates, we can then construct a band around the empirical CDFs using the sharp constant \citep{Massart-1990-DKW-Tight-Constant} of the distribution-free DKW inequality \citep{DKW-1956-DKW-Inequality}, calculating the error for each class $c \in \{1, \ldots, C \}$ from the corresponding index in $\mathbf{\hat{n}}$ if $\hat{n}_c > 0$:
\begin{align}\label{equation:dkw} 
\epsilon_{c} = \sqrt{\frac{1}{2 \cdot \hat{n}_c } \log_e \left ( \frac{2}{1-\alpha'} \right )}
\end{align}
If $\hat{n}_c = 0$, our convention is to set $\epsilon_{c}=1$. We can then construct the lower and upper counterparts to the quantile vector of Eq.~\ref{equation:sdm-output-cdfs}:
\begin{align}\label{equation:sdm-output-cdfs-lower} 
\vv_{\rm{lower}} =  [ 
&\min \left (
\max \left ( 
	{\rm{eCDF}}^{y_1}_{\rm{ca}}( \sdm(\vz')_1 ) - \mathbf{1}_{\hat{y}=1} \cdot \epsilon_{1} + \mathbf{1}_{\hat{y}\ne 1} \cdot \epsilon_{1}, 
0 \right ), 
1 \right ),
 \ldots , \notag \\ 
& \min \left (
\max \left ( 
	{\rm{eCDF}}^{y_C}_{\rm{ca}}( \sdm(\vz')_C ) - \mathbf{1}_{\hat{y}=C} \cdot \epsilon_{C} + \mathbf{1}_{\hat{y}\ne C} \cdot \epsilon_{C}, 
0 \right ), 
1 \right )
]
\end{align}
\begin{align}\label{equation:sdm-output-cdfs-upper} 
\vv_{\rm{upper}} =  [ 
&\min \left (
\max \left ( 
	{\rm{eCDF}}^{y_1}_{\rm{ca}}( \sdm(\vz')_1 ) + \mathbf{1}_{\hat{y}=1} \cdot \epsilon_{1} - \mathbf{1}_{\hat{y}\ne 1} \cdot \epsilon_{1}, 
0 \right ), 
1 \right ),
 \ldots , \notag \\ 
& \min \left (
\max \left ( 
	{\rm{eCDF}}^{y_C}_{\rm{ca}}( \sdm(\vz')_C ) + \mathbf{1}_{\hat{y}=C} \cdot \epsilon_{C} - \mathbf{1}_{\hat{y}\ne C} \cdot \epsilon_{C}, 
0 \right ), 
1 \right )
]
\end{align}
from which $\softqbinLower$ and $\softqbinUpper$ follow:
\begin{align}\label{equation:softqbin-lower-upper} 
\softqbinLower &= log_{e} \left ( (2+q)^{\vv_{{\rm{lower}}_{\hat{y}}}} \right ) \\
\softqbinUpper &= log_{e} \left ( (2+q)^{\vv_{{\rm{upper}}_{\hat{y}}}} \right )
\end{align}
Analogous to Eq.~\ref{equation:rescaled-prediction-conditional-estimate}, we then construct our estimators after rescaling $\vv_{\rm{lower}}'  = \mW^{''T} \vv_{\rm{lower}}$, $\vv_{\rm{lower}}' \in \reals^C$ and $\vv_{\rm{upper}}'  = \mW^{''T} \vv_{\rm{upper}}$, $\vv_{\rm{upper}}' \in \reals^C$:

\begin{align} 
\pLower &= \frac{
(2+\softqbinLower)^{~v'_{{\rm{lower}}_{\hat{y}}} }
}{
\sum^C_{c=1}{(2+\softqbinLower)^{~v'_{{\rm{lower}}_{c}} }}
} \label{equation:rescaled-prediction-conditional-estimate-lower} \\ 
\pCentroid &= o_{\hat{y}} ~~~\rhd \text{from Eq.~\ref{equation:rescaled-prediction-conditional-estimate}} \\
\pUpper &= \frac{
(2+\softqbinUpper)^{~v'_{{\rm{upper}}_{\hat{y}}} }
}{
\sum^C_{c=1}{(2+\softqbinUpper)^{~v'_{{\rm{upper}}_{c}} }}
} \label{equation:rescaled-prediction-conditional-estimate-upper}
\end{align}
As with Eq.~\ref{equation:rescaled-prediction-conditional-estimate}, the convention is to set $\softqbinLower = 0$ and/or $\softqbinUpper = 0$ for the rare cases for which the transforms in Eq.~\ref{equation:rescaled-prediction-conditional-estimate-lower} and/or Eq.~\ref{equation:rescaled-prediction-conditional-estimate-upper}, respectively, result in the $\argmax$ value of the normalized output vector not being equivalent to $\hat{y} = \argmax \vz'$. (In such cases, e.g., Eq.~\ref{equation:rescaled-prediction-conditional-estimate-lower} is not re-calculated with $\softqbinLower = 0$, but rather such values are treated separately in downstream analyses as out-of-distribution points.)
\paragraph{Base Estimators.} $\pLower \in \reals^1$ will be used as the basis of our primary test-time estimator of prediction-conditional uncertainty (see \S~\ref{sec:test-time-estimates} for the complete, index-conditional estimator). $\pCentroid \in \reals^1$ (via Eq.~\ref{equation:rescaled-prediction-conditional-estimate}) is a consequence of intermediate results needed in service of constructing $\pLower$ (e.g., for training the re-scaler and setting a threshold on $\softqbin$, described below), whereas $\pUpper \in \reals^1$ is primarily only of research interest, included here to analyze the behavior of the approach.\footnote{In practice, rather than using $\pUpper$, if a less stringent admission criteria is desired, the operative action is to reduce $\alpha'$ and re-estimate $\pLower$.} 

\subsubsection{Training the Rescaling Transform}\label{sec:training-the-rescaler}
We train the $C^2$ parameters of $\mW^{''}$ of the re-scaling linear layer over $\calibrationSplit$ (\textit{not} $\trainSplit$) by minimizing the following loss (Alg.~\ref{alg:train-rescaler}), which is the counterpart to Eq.~\ref{equation:rescaled-prediction-conditional-estimate}, while all other parameters remain fixed:
\begin{align}\label{eq:rescaler-loss}
\gL(\mW^{''}; \calibrationSplit) = -\frac{1}{| \calibrationSplit |} \sum_{n}^{| \calibrationSplit |} \log_{(2+\softqbin)}\left(
\frac{
(2+\softqbin)^{v'_{y_n}}
}{
\sum^C_{c=1}{(2+\softqbin)^{v'_c}}
}
\right)
\end{align}
Our convention is to train with a batch size of 1 and conclude the learning process if $\gL(\mW^{''}; \calibrationSplit)$ increases for a pre-specified (as a hyper-parameter) number of consecutive epochs.

\begin{algorithm}
    \caption{Training the Weights of the Rescaling Transform}
    \label{alg:train-rescaler}
    \small
    \begin{algorithmic}[1] 
        \Require cached $\vv$ for $\calibrationSplit$, $\text{rescaler max epochs}$, $\text{rescaler stopping condition}$ %
        \Procedure{train-rescaler}{cached $\vv$ for $\calibrationSplit$, rescaler max epochs, rescaler stopping condition}
        		\State $\mW^{''}_{*} \gets \emptyset$ \Comment{Final weights}
        		\State $\mW^{''} \gets$ random initialization
		\State $\rm{metric} \gets \infty$
		\State $\rm{counter} \gets 0$
             	\For{$e \in 1, \ldots, \text{rescaler max epochs}$}
		 	\State Minimize ${\rm{loss}} \gets \gL(\mW^{''}; \calibrationSplit)$ \Comment{Eq.~\ref{eq:rescaler-loss}}
          	 	\If{${\rm{loss}} < \rm{metric}$}
           			\State $\rm{metric} \gets {\rm{loss}}$
				\State $\mW^{''}_{*} \gets \mW^{''}$
           		\EndIf
		        \If{${\rm{loss}} > \rm{metric}$}
           			\State $\rm{counter} \gets \rm{counter} + 1$
				\If{$\rm{counter} > \text{rescaler stopping condition}$}
					\State \textbf{break} 
				\EndIf
			\Else
				\State $\rm{counter} \gets 0$
           		\EndIf
		 \EndFor
             \State \textbf{return} $\mW^{''}_{*}$
        \EndProcedure
        \Ensure $\mW^{''}_{*}$
    \end{algorithmic}
\end{algorithm}

\subsubsection{Region-specific eCDFs}\label{sec:region-specific-ecdfs}

The estimator $\pLower$ incorporates an explicit notion of the effective sample size. Smaller effective sample sizes will be associated with lower probability estimates (and vice-versa). It also has a strong relative notion of the highest probability regions of the output distribution by virtue of the original $\Similarity$, $\Distance$, and $\Magnitude$ signals, and the aggregated distributional statistics over these signals. However, it lacks a human interpretable, principled cutoff, or threshold, by which we can have some assurance that the new points we see are reasonably comparable to the data we observed in deriving our estimator. This is a more subtle and foundational problem than it may initially seem; we must account for distribution shifts if we seek to realistically achieve our desired notion of index-conditional calibration (Def.~\ref{defn:index-conditional-calibration}). It will require an additional set of transforms to resolve, even with the already strong signals of prediction-conditional uncertainty from our estimator, to which we turn next.

It follows from Eq.~\ref{eq:softmax} that the output of $\softmax(\vz)$ can be viewed as $\softmax(\vz)= \triangle^{C-1}$, which is the ($C-1$)-dimension simplex, where the dimension reduction is a consequence of the output summing to 1. The same is true of the normalized value $\vo$. If we instead consider the over-parameterized version in which each event probability of the categorical distribution (e.g., Eq.~\ref{eq:softmaxCategorical}) is explicitly specified as an element of a vector of length $C$, the following indicator result directly follows:
\begin{remark}\label{class-conditional-thresholding}
Given the $C$ class-conditional CDFs over categorical distributions where the $1-\alpha'$ $(\alpha' \in (\frac{1}{C},1])$ quantile threshold $\psi_c$ $(\psi_c \in [0,1])$ of each class $c \in \{1,\ldots,C\}$ is $> \frac{1}{C}$ (i.e., $\psi_c={\rm{inverseCDF}}^{y_c}(1-\alpha') > \frac{1}{C} ~\forall~c~\in \{1,\ldots,C\}$), a set of i.i.d. points sampled from the same distribution as the CDFs, each of whose event probability vector $\ve = [e_1, \ldots e_C]$ has one (1) element at least the corresponding class threshold (i.e., $\left |[e_1, \ldots e_C] \ge [\psi_1, \ldots \psi_C] \right | = 1$, with the comparison taken element-wise), will have class-conditional accuracies $\ge \alpha'$, in expectation.
\end{remark}
\begin{proof}
Partition the class-conditional CDFs of the categorical distributions, for which 
$\psi_c={\rm{inverseCDF}}^{y_c}(1-\alpha') > \frac{1}{C} ~\forall~c~\in \{1,\ldots,C\}$, at $[\psi_1, \ldots \psi_C]$.
 The resulting high-probability partitions---those $\ge \psi_c$---are $C$ Bernoulli distributions each with success probability $p_c \ge \alpha'$. Take as $[n_1, \ldots n_C]$ the class-wise count of i.i.d. points whose event probability vector, $\ve$, satisfies $\left |[e_1, \ldots e_C] \ge [\psi_1, \ldots \psi_C] \right | = 1$. Then by the definition of the expected value of a Binomial distributed random variable, it follows from these trials that $[\frac{n_1 \cdot p_c}{n_1}, \ldots, \frac{n_C \cdot p_C}{n_C}] = [\ge \alpha', \ldots, \ge \alpha']$, which is the desired class-conditional accuracy for this restricted set of points. Now, instead assume that one or more of the Bernoulli distributions has a success probability $p_c < \alpha'$. This implies that the class-conditional CDFs were constructed from a distribution whose event probabilities are not those of the $(C-1)$-dimension simplex since we require $\psi_c={\rm{inverseCDF}}^{y_c}(1-\alpha') > \frac{1}{C} ~\forall~c~\in \{1,\ldots,C\}$ with the CDFs constructed class-wise relative to the true labels, which is a contradiction of the definition of a categorical distribution since the sum of all event probabilities, each of which is a real value in $[0,1]$, must equal 1.
\end{proof}
Note that when $\psi_c < \frac{1}{C}$ no such assurance across all classes necessarily results, since the resulting thresholding of the probability vectors may induce a complex dependence across the class-conditional CDFs.\footnote{We leave to future work whether a similar result holds for a subset of the class-conditional accuracies if only \textit{some}, rather than \textit{all}, class-wise thresholds are at least $\frac{1}{C}$. In our primary verification setting over LLMs, the typical setting is $C=2$, or some similarly small $C \in \sZ^{2+}$, where to ensure deployment reliability, the LLM would not be deployed until the verification task yields class-conditional accuracies for all classes at or above $\alpha'$ on the available labeled sets, so we do not consider this case here.} In such cases, the thresholding of a new point may result in multiple classes above the threshold, and the subsequent stratification of this set of points to those for which $\left |[e_1, \ldots e_C] \ge [\psi_1, \ldots \psi_C] \right | = 1$ will not necessarily have class-conditional accuracies $\ge \alpha'$, in expectation.

Remark~\ref{class-conditional-thresholding} thus differs from set-valued estimators such as conformal estimators \citep{Vovk-2005-AlgorithmicLearningBook}, which as previously mentioned (see \S~\ref{sec:sdm-calibration}, introduction) are premised on a different calibration compromise. For example, with conformal estimators, there is a statistical assurance for coverage of the true class in a discrete prediction set (itself a distinct quantity from that considered here) across all points regardless of the distribution of the conformity score (e.g., instead of a categorical distribution, a conformity score can be an unnormalized scoring function), but no assurance conditional on the subset of high-probability points. We explore the implications of these tradeoffs in our empirical experiments.

Remark~\ref{class-conditional-thresholding} can be viewed as a useful indicator function, but it is not particularly informative as an estimator alone. We will use it in service of dividing the output distribution into high probability regions via $\softqbin$, described next.

\paragraph{Corralling the high-probability region via exclusion of the observed high-epistemic-uncertainty points.}
Intuitively, higher values of $\softqbin$ correspond to points with a closer connection to the observed data and thus lower epistemic uncertainty, as this single value takes into account the $\Similarity$, $\Distance$, and $\Magnitude$ signals, and distributional statistics over those signals. The result in Remark~\ref{class-conditional-thresholding} provides a principled basis for setting a threshold on $\softqbin$ over $\calibrationSplit$ that we can then apply at test time, without access to the true label, to constrain our estimates to the high-probability region of the distribution.

The value of $\softqbin$ is real-valued, but only $\le | \calibrationSplit |$ values are observed, so a simple iterative search algorithm is sufficient to find the value of $\softqbin$ that satisfies Remark~\ref{class-conditional-thresholding} such that all thresholds, $\psi_c$, over the estimates of $\vo$ (Eq.~\ref{equation:rescaled-prediction-conditional-estimate}), are at least $\alpha'$. By definition, $\alpha' > \frac{1}{C}$, so this more stringent requirement satisfies the condition in Remark~\ref{class-conditional-thresholding}, while also requiring $\softqbin$ to be restricted to the prediction-conditional estimates of $\pCentroid \ge \alpha'$. The full algorithm appears in Alg.~\ref{alg:estimate-soft-q-bin}, iteratively constructing class-wise eCDFs over $\calibrationSplit$ restricted to progressively larger values of $\softqbin$. (These eCDFs over the $\vo$ values of $\calibrationSplit$ are only needed for Alg.~\ref{alg:estimate-soft-q-bin} and are not needed at test time, unlike those of Eq.~\ref{equation:d}, Eq.~\ref{equation:sdm-output-cdfs}, and Eq.~\ref{equation:effective-sample-size}.) Note that we only consider values of $\hardqbin > 0$, as points with $\hardqbin = 0$ are considered out-of-distribution.\footnote{The reason for the floor operation becomes evident in the next section. $\hardqbin$ will serve as our hard-partitioned mapping between the observed data and new test points to enable estimates of uncertainty over iterations of the entire process described thus far.} The search algorithm may fail to find a suitable final value, $\minValidQBin$, at which point the operative conclusion is that reliable estimates of index-conditional calibration (Def.~\ref{defn:index-conditional-calibration}) are not possible without reducing $\alpha'$, or acquiring additional data and/or a stronger model.\footnote{Alg.~\ref{alg:estimate-soft-q-bin} could be readily modified to find an adaptive value of $\alpha'$, iteratively reducing $\alpha'$ if a suitable $\minValidQBin$ value is not found. However, in practice, determining $\alpha'$ for LLM settings is a decision made exogenous to the model development process. We seek to develop our models (and data) to meet a given $\alpha'$ value, rather than the other way around, so we do not consider that variation here.}

When a value of $\minValidQBin$ can be found, the convention is to restrict our estimates of index-conditional calibration to the new, unseen test points that satisfy $\softqbinLower \ge \minValidQBin$ after considering the final additional sources of uncertainty from the data splitting and learning processes, which we consider next.

\begin{algorithm}
    \caption{Search Algorithm to Find $\minValidQBin$ to Detect High-Probability Regions}
    \label{alg:estimate-soft-q-bin}
    \small
    \begin{algorithmic}[1] 
        \Require cached $(\softqbin, \vo)$ for $\calibrationSplit$, $\alpha' \in (\frac{1}{C},1]$ %
        \Procedure{find-min-rescaled-q}{cached $(\softqbin, \vo)$ for $\calibrationSplit$, $\alpha' \in (\frac{1}{C},1]$}
        \State $\minValidQBin \gets \emptyset$ \Comment{A suitable $\minValidQBin$ may not exist.}
        \State $ [\psi_1, \ldots \psi_C] \gets [\emptyset, \ldots, \emptyset]$ \Comment{Needed at test-time, if applicable}
        \State $\tilde{q}s \gets {\rm{sorted}} ~[\tilde{q} \in \calibrationSplit~{\rm{s.t.}}~\hardqbin > 0]$ \label{line:ood-hardqbin} \Comment{Restricted to $\hardqbin > 0$ to exclude OOD}
        \For{$\softqbin' \in \tilde{q}s$} 
            \State Construct ${\rm{eCDF}}^{y_1}_{\rm{ca}} , \ldots , {\rm{eCDF}}^{y_C}_{\rm{ca}} $ for all $ \softqbin \ge \softqbin'$ in $\calibrationSplit$\label{line:eCDFs} \Comment{eCDFs for $\vo$ (Eq.~\ref{equation:rescaled-prediction-conditional-estimate}), stratified by $y$}
            \State Calculate $\psi_c={\rm{inverseCDF}}^{y_c}_{\rm{ca}}(1-\alpha')  ~\forall~c~\in \{1,\ldots,C\}$ \Comment{Quantile functions are inverses of L.~\ref{line:eCDFs}}
	\If{${\rm{all}}( ~ [\psi_1, \ldots \psi_C] \ge \alpha' ~)$} \Comment{Element-wise comparison}
        	    \State $\minValidQBin \gets \softqbin'$ \Comment{Satisfies Remark~\ref{class-conditional-thresholding} at the prediction-conditional estimate (see text) of $\ge \alpha'$}
	    \State \textbf{break}
        \EndIf        
        \EndFor
        \State \textbf{return} $\minValidQBin$, $ [\psi_1, \ldots \psi_C]$
        \EndProcedure
        \Ensure $\minValidQBin$, $ [\psi_1, \ldots \psi_C]$
    \end{algorithmic}
\end{algorithm}

\subsubsection{Accounting for Uncertainty in the Data Splitting and Learning Processes}\label{sec:iterative-uncertainty}

As a final step, we take into account uncertainty over the data splitting and learning processes. This will incur non-trivial additional computational costs, but these are one-time development costs for an estimator. At test time, our estimates will be constant offsets on $\minValidQBin$ and $\pLower$, the latter conditional on $\hardqbin \in \sZ^{0+}$, which will serve as a stable mapping between $\calibrationSplit$ and new, unseen test points. In summary, in this section, we seek:
\begin{align}\label{eq:iteratedQuantities}
\CauchyMinValidQBin \hspace{2em} &\rhd \text{A robust estimate of $\minValidQBin$} \\ 
\CauchyMagnitudeIteratedLowerEstimate \hspace{2em} &\rhd \text{A class-wise, robust correction for $\pLower$, conditional on $\hardqbin$}
\end{align}

Conceptually, the estimation process is straightforward. We repeat the training and estimation processes described above $J$ times and derive our constant offsets via summary statistics over those estimates. The one complication that arises is that we will have to depart from the distribution-assumption-light approaches above, since $J$ will typically not be large due to the computational expense. (The full process across $J$ iterations to construct a single estimator needs to remain reasonably computationally lightweight relative to an LLM training epoch, as it itself will be embedded into the training loop of an LLM, described below.) Instead, we will estimate each of these processes as a Cauchy distribution, given its relatively wide tails and relatively robust scale parameter. 

A Cauchy distribution is defined by a location parameter, $\CauchyDistributionLocation$, and a scale parameter, $\CauchyDistributionScale$:
\begin{align}\label{eq:cauchyDistribution}
\CauchyDistribution(\CauchyDistributionLocation, \CauchyDistributionScale)
\end{align}
The inverse CDF (i.e., quantile function) of a Cauchy distribution for a particular quantile, $\alpha \in [0,1]$, can be calculated analytically as:
\begin{align}\label{eq:cauchyDistributionInverseCDF}
{\rm{inverseCDF}}_{\CauchyDistribution(\CauchyDistributionLocation, \CauchyDistributionScale)}(\alpha) = \CauchyDistributionLocation + \CauchyDistributionScale \tan \left ( \pi \left (\alpha - \frac{1}{2} \right ) \right )
\end{align}
We take as our estimate of $\CauchyDistributionScale$ the median absolute deviation around the median of our sample ($\CauchyMAD$).

\paragraph{Robust detection of high-probability regions.} To calculate $\CauchyDistributionScale$ for $\CauchyMinValidQBin \in \reals^1$, we take the $\CauchyMAD$ of the $J$ estimates of $\minValidQBin$. The location parameter is taken as $\minValidQBin_*$, the estimate of $\minValidQBin$ over the model with the final chosen weights (see Alg.~\ref{alg:sdm-estimator-training}). We can then analytically calculate our desired value via Eq.~\ref{eq:cauchyDistributionInverseCDF} at $\alpha' \in (\frac{1}{C}, 1]$:
\begin{align}\label{eq:cauchyMinValidQBin}
\CauchyMinValidQBin = {\rm{inverseCDF}}_{\CauchyDistribution(\minValidQBin_*, \CauchyDistributionScale)}(\alpha')
\end{align}
Note that since $\alpha'$ corresponds to the right-tail of the distribution, $\CauchyMinValidQBin \ge \minValidQBin_*$, i.e., a more restrictive threshold on the high-probability region. In scenarios (not considered in the experiments here) where the computational budget necessitates $J=1$, the convention would be to take $\CauchyMinValidQBin \coloneq \minValidQBin_*$, with a tacit assumption that these additional sources of uncertainty have not been explicitly accounted for.

\paragraph{Robust output adjustment.} To calculate $\CauchyDistributionScale$ for $\CauchyMagnitudeIteratedLowerEstimate$ conditional on $\hat{y}$ and $\hardqbin$ (i.e., $\CauchyDistributionScale \given \hat{y}, \hardqbin$), we take the $\CauchyMAD$ of the $J$ \textit{medians} (as written) of $\pCentroid$ over $\calibrationSplit$, conditional on $\hat{y}$ and $\hardqbin$.\footnote{Implementation note: Unlike the eCDFs, which are constructed by stratifying on the true label, $y$, in $\calibrationSplit$, this quantity is calculated by stratifying on the predicted label, $\hat{y}$, in $\calibrationSplit$.} Similar to above, we can then calculate:
\begin{align}\label{eq:cauchyIteratedLower}
\CauchyMagnitudeIteratedLowerEstimate =  {\rm{inverseCDF}}_{\CauchyDistribution(0, (\CauchyDistributionScale \given \hat{y}, \hardqbin ) )}(\alpha')
\end{align}
In this case, $\CauchyDistributionLocation$ is 0, as $\CauchyMagnitudeIteratedLowerEstimate$ will be subtracted from $\pLower$ as an offset, an assumption that each distribution is centered on the given point. To simplify the presentation (and since the upper offset is not needed in practice), we only consider this as a lower offset on our base estimators.

As $\hardqbin$ increases, the number of points in the sample will tend to decrease, but so will the $\CauchyMAD$, so the estimates remain reasonable in practice. As we will see in our experiments, high values of $\hardqbin$ (that are otherwise attested in $\calibrationSplit$) are not uncommonly associated with $\CauchyMAD$ values that are within 0 of numerical precision.

As with $\CauchyMinValidQBin$, although it is generally recommend to take these additional sources of uncertainty into consideration, when $J=1$, the convention would be to take $\CauchyMagnitudeIteratedLowerEstimate \coloneq 0$.
                    
\subsubsection{Index-Conditional Calibration}\label{sec:test-time-estimates}

With the above models and estimators, we can now robustly calculate the index-conditional uncertainty of a new, unseen test point $\vx \in \testSplit$.

We first take as the prediction $\hat{y} = \argmax{\vz'}$. Then, with $\trainSplit$ to calculate $\q$ and $\dNearest$; the cached class-wise empirical CDFs over $\calibrationSplit$ of Eq.~\ref{equation:d}, Eq.~\ref{equation:sdm-output-cdfs}, and Eq.~\ref{equation:effective-sample-size}; $\CauchyMinValidQBin$ and the thresholds ($ [\psi_1, \ldots \psi_C]$); and $\CauchyMagnitudeIteratedLowerEstimate$, the index-conditional uncertainty estimate of $p(y \given \vx)$ at $\alpha'$ (Def.~\ref{defn:index-conditional-calibration}) is: 

\begin{align}\label{eq:index-conditional-sdm-estimator}
\EstimatorPLower = 
\begin{cases}
  \max(0, \pLower - \CauchyMagnitudeIteratedLowerEstimate[\hat{y}]) & \text{if~} \left [ \softqbinLower \ge \CauchyMinValidQBin \right ] \wedge \left [ \left ( \pLower - \CauchyMagnitudeIteratedLowerEstimate[\hat{y}] \right ) \ge \psi_{\hat{y}} \right ] \\
  \bot & \text{otherwise}
\end{cases}
\end{align}
where $\bot$ indicates a rejected (non-admitted) point.\footnote{At test time, the mapping to the $\hardqbin$-conditioned statistics (i.e., $\CauchyMagnitudeIteratedLowerEstimate[\hat{y}]$) is via $\hardqbinLower$ for the test instance.}

As noted in the previous sections, in the rare cases when the transforms after the $\sdm$ activation result in the $\argmax$ index not matching $\hat{y}$, we set $\softqbinLower = 0$, which effectively treats the point as out-of-distribution. In such cases, $\EstimatorPLower = \bot$, since $\CauchyMinValidQBin > 0$ as a consequence of Line~\ref{line:ood-hardqbin} in Alg.~\ref{alg:estimate-soft-q-bin}. 

Our convention in subsequent sections will be to refer to summary statistics and comparisons of $\EstimatorPLower$ (Eq.~\ref{eq:index-conditional-sdm-estimator}), excluding the points assigned $\bot$, as estimates from the ``estimator $\EstimatorPLower$''. We do the same for the ``estimator $\EstimatorPCentroid$'' and the ``estimator $\EstimatorPUpper$'', but where the latter two quantities are calculated from the corresponding centroid and upper intermediate quantities, respectively.

\paragraph{Complexity.} The added computational overhead over an underlying $\underlyingNetwork$ with a $\softmax$ activation is dominated by calculating $\q$ (and by extension, $\dNearest$). The transforms after the $\sdm$ activation function add negligible additional overhead. For perspective, this is on the order of the additional computation needed for commonly used dense retrieval augmentations of LLMs, so it is readily achievable at interactive speeds in practice.

\paragraph{Sharpness.} As noted in \S~\ref{sec:sdm-calibration}, we seek estimators that are both informative (i.e., not unnecessarily rejecting correct predictions) and robust (i.e., we prefer rejection over falling under the expected $\alpha'$ accuracy). The above transforms seek to achieve this by taking the uncertainty signals from an $\sdm$ activation and further separating the high and low probability regions of the distribution, as well as providing a hard cut via $\CauchyMinValidQBin$ to altogether exclude predictions over high epistemic uncertainty regions. We explore these behaviors empirically in our experiments.

Next, we incorporate our estimators directly into LLM next-token training. 

\subsection{From SDM Calibration to SDM Networks}\label{sec:sdm-networks}
The above approach is already a very powerful and easily implemented mechanism for building complex LLM pipelines. We can treat an underlying $\underlyingNetwork$ as fixed, add an $\sdm$ activation layer, and then use the $\sdm$ estimator for conditional branching for test-time compute, retrieval, tool-calling, and related.

However, earlier in the model development pipeline (e.g., as done by LLM model providers), we need a mechanism for fine-tuning a $\underlyingNetwork$ after the initial unsupervised training stage.\footnote{In principle, the methods in this section can also be used for bootstrapping a randomly initialized LLM against an existing (possibly larger) model, which we leave to future work.} In this section, we show how to incorporate the $\sdm$ mechanism directly into the LLM next-token training process. We will refer to this process and the resulting model as an $\sdm$ network.

Conceptually, an $\sdm$ activation and estimator over an averaged history of frozen hidden states and the token-level hidden state will be trained for binary classification at the unit of analysis of the available labels (e.g., the document-level). This estimator then provides the $\Similarity$ and $\Distance$ values for an $\sdm$ activation for next-token loss of the LLM during training. Because an $\sdm$ activation does not alter the $\argmax$ prediction, greedy token-level generation can proceed without the computational cost of the $\sdm$ activation at every token at test time, with the global $\sdm$ estimator providing verification over the final generation. \textbf{This process shares the same goal of existing fine-tuning approaches to increase overall accuracy, add information to a model, etc., as well as the new goal of increasing the proportion of verifiable high-probability generations from a model.} During training, we seek to penalize the model for verification mistakes, and reward the model for increasing the cardinality of the set of admitted points.  

We first introduce our data encoding scheme in \S~\ref{sec:verification-encoding} for verification. Next, orthogonal to the $\sdm$ mechanism itself, we introduce a parsimonious regularization method (\S~\ref{sec:negative+positive}) to enable fine-tuning on a small amount of data while discouraging catastrophic forgetting. Finally, we introduce the process for training the $\sdm$ network (\S~\ref{sec:sdm-network-details}).

\subsubsection{Universal Verification Encoding}\label{sec:verification-encoding}
In the abstract, our data is similar to that in the previous sections: Input documents accompanied with discrete labels. However, while we previously treated each document, $\vx$, as a single atomic unit, we will now also be concerned with the individual tokens of the document, for which we use the notation $\trainSplit=\{(\vx_n = [x_1, \ldots, x_T], y_n, [y_n^{\rm{task}}])\}_{n=1}^{N}$ for our labeled training set, and similarly for our labeled calibration set, $\calibrationSplit$. Each token, $x_t \in \{1, \ldots, | \gV | \cdot 2 \}$, is represented as an index into a vocabulary, where $\gV$ is the vocabulary of the LLM trained during the initial unsupervised training stage. The reason for the factor of 2 is described in the next section. Implicit in our representation is that each instance will have a marker at some $x_t$ indicating a ``completion'' (i.e., a sequence after an instruction prompt or prefix, more generally). Our document-level labels, $y \in \gY = \{0, 1\}$, are as in previous sections, but specifically restricted to binary classification, where the convention is to treat $y=0$ as representing the \texttt{unverified} class and $y=1$ as the \texttt{verified} class (i.e., an acceptable generation, conditional on the instruction or context).

For some documents, we have classification labels, $\yTask \in \sZ^{2+}$, for the underlying tasks encoded in the data. For example, for a sentiment classification task of negative and positive reviews, $y=0$ for verification when the classification decision is wrong, whereas $y=1$ for verification when the classification decision is correct. Among those for $y=1$, $\yTask=0$ could indicate a negative review and $\yTask=1$ could indicate a positive review. These task-specific labels are predicted via the generated text of the LLM, and if available, we can use them during training (e.g., as part of our stopping criteria to choose the best weights, by parsing the generated text and comparing to $\yTask$). Unlike typical classification settings, these labels may---and typically will---cover multiple disparate tasks; hence, the designation of \textit{universal} verification. When the distinction is potentially ambiguous, we will add a superscript to $y$ for the binary verification labels: $\yVerification$. 

Unlike typical preference fine-tuning encodings, we do not require prefixes (or prompts) of $\vx$ to be paired with different completions and opposing document-level labels. However, as in the above sections, we will assume that $\trainSplit$ and $\calibrationSplit$ are balanced across $y$ (i.e., an approximately equal number of documents with $y=0$ and $y=1$).

The $\sdm$ activation layer for verification will be trained with $\trainSplit$, seeing all documents with $y=0$ and $y=1$ labels. However, the LLM's $\sdm$ activation for next-token training will only directly see documents with $y=1$, with the signal of the \texttt{unverified} class coming indirectly via matching into $\trainSplit$ to calculate $\Similarity$ and $\Distance$. (There is an additional nuance with train-time generation vs. train-time force-decoding that will be clarified below.) As such, the additional $\sdm$ mechanisms enable a unification of preference fine-tuning, instruction fine-tuning, and supervised fine-tuning encodings since in all of the above, we always have at least the $y=1$ documents, and it is typically straightforward to collect, or otherwise synthetically generate, unpaired examples to serve as $y=0$ (i.e., generations we seek to avoid showing users).

\subsubsection{Negative+Positive Vocabulary Normalization and Regularization}\label{sec:negative+positive}
Before we can make progress on incorporating the $\sdm$ mechanisms, we need to address the matter of fine-tuning pre-trained LLMs without inducing catastrophic forgetting. This is critical, since each round of LLM training and fine-tuning is computationally expensive. We seek to make incremental changes to the model without having to run subsequent learning processes over all previously seen data. To address this, we first briefly recall the training of auto-regressive neural language models prior to the era of large-scale pre-training.

\paragraph{[L]LM Training Redux.} Prior to the era of large-scale pre-training of LMs that emerged at the end of the 2010's, auto-regressive language models for transduction tasks (e.g., grammatical error correction) were successfully trained from random initialization using specialized input control tokens and output diff sequences (and associated output control tokens) that separated non-preferred (pre-transduction) and preferred (post-transduction) generated sequences \citep{Schmaltz-Etal-2017-DiffSequenceTransduction}.
Importantly, the bias on the diff control sequences could be modulated to control precision and recall over the absence and presence of the transduction operation \citep{Schmaltz-Etal-2016-ClassificationAsTransduction}. In-effect, the sequence transduction model could be effectively used as a classifier without additional classification layers, while also having the expressivity to generate token sequences, unlike standard discrete classifiers.

Input and output control tokens are now prominent features of LLM vocabularies to structure prompts, instructions, and reasoning sequences. However, while the bias of individual tokens can be modified with an additive offset, current LLMs lack a mechanism to explicitly bifurcate the output distribution into non-preferred and preferred regions in the manner of the earlier models. This capability can be (re)-added to LLMs without direct training on diff transduction sequences, as follows.

\paragraph{Negative+Positive Vocabulary Normalization.}

Consider a pre-trained LLM model, $\gM^{\rm{ref}}$. Our reference model generates acceptable sequences over part of the data distribution, but it also produces non-preferred (\textsc{negative}) generations; hence, our desire for further training. However, we only want to alter the behavior of $\gM^{\rm{ref}}$ over the space that produces \textsc{negative} generations, otherwise we may unexpectedly cause the previously acceptable space of generations to also become \textsc{negative}. In effect, we have two regions---a bifurcation---of the output distribution: The space of existing acceptable generations and the space of \textsc{negative} generations. We seek to replace the \textsc{negative} region with a new \textsc{positive} region of acceptable generations without (or at least minimally) impacting the existing acceptable region.

From $\gM^{\rm{ref}}$ create two clones, $\gM^{\rm{neg}}$ and $\gM^{\rm{pos}}$. Each model has a final linear layer that maps to the output vocabulary, $\gV$, via a weight matrix\footnote{We ignore the bias terms here to simplify the presentation. In practice, it is not uncommon for $\vb=\mathbf{0}$.}: $\vz_{\rm{ref}} = \mW^T_{\rm{ref}} \vh_{\rm{ref}}$, $\vz_{\rm{neg}} = \mW^T_{\rm{neg}} \vh_{\rm{neg}}$, and $\vz_{\rm{pos}} = \mW^T_{\rm{pos}} \vh_{\rm{pos}}$, respectively. During fine-tuning for the next-token loss, we then calculate the $\sdm$ activation (in-place of a standard $\softmax$) as the concatenation of the un-normalized output of $\gM^{\rm{neg}}$ and $\gM^{\rm{pos}}$, $\sdm(\vz_{\rm{neg}}, \vz_{\rm{pos}})$, keeping the weights of $\gM^{\rm{neg}}$, $\mW_{\rm{neg}}$ and $\theta_{\rm{neg}}$, fixed and updating the weights of $\gM^{\rm{pos}}$, $\mW_{\rm{pos}}$ and $\theta_{\rm{pos}}$. For the $y=1$ documents that participate in fine-tuning, we simply take the original token indexes and add an offset, $x_t + |\gV|$, for the output tokens when calculating the loss over the joint, concatenated distribution. (Input tokens retain their original indexes.) At test time, the $\argmax$ output $\text{index}~{\rm{mod}}~|\gV|$ maps back to the original token symbol in the vocabulary. In this way, an additional set of token symbols is never explicitly instantiated.

In the most direct sense, this then requires a copy of the full weights to be present at test time. However, in practice, $\gM^{\rm{pos}}$ need not be a copy of all the weights; $\gM^{\rm{pos}}$ can be represented by adaptor layers, or similar mechanisms (e.g., only updating a subset of the model's weights). 

\paragraph{Regularization.} To further prevent drift from the original reference distribution, we also add an $L^2$ regularization term in the $\log_{(2+q)}$ space of the normalized joint, concatenated distribution when calculating the next-token loss:
\begin{align}
\regularizationTerm = || \vi \odot \log_{(2+q)} \left ( \sdm(\vz_{\rm{ref}}, \vz_{\rm{ref}}) \right ) -  \vi  \odot \log_{(2+q)} \left ( \sdm(\vz_{\rm{neg}}, \vz_{\rm{pos}}) \right ) ||_2
\end{align}
where the Hadamard (element-wise) product ($\odot$) is with a mask vector $ \vi \in \reals^{|\gV| \cdot 2}$ that lessens the regularization on the peak of the distribution by not considering the $\argmax$ indexes of the reference, negative, and positive distributions, as well as that of the ground-truth next-token label (here, represented as $t$), in the $L^2$ constraint:\footnote{This also accounts for the setting, not considered in our experiments, where $\gM^{\rm{neg}}$ is not identical to $\gM^{\rm{ref}}$, such as via multiple iterations of fine-tuning where $\gM^{\rm{neg}}$ is trained away and is replaced with $\gM^{\rm{pos}}$ for a subsequent fine-tuning round.}
\begin{align}\label{eq:regularization-mask}
\vi &= \mathbf{1} \in \reals^{|\gV| \cdot 2} \\
\vi_{\argmax { (\vz_{\rm{ref}})}} &= 0 \notag \\
\vi_{\argmax { (\vz_{\rm{ref}})} + |\gV| } &= 0 \notag \\
\vi_{\argmax { (\vz_{\rm{neg}})}} &= 0 \notag \\
\vi_{\argmax { (\vz_{\rm{pos}})} + |\gV| } &= 0 \notag \\
\vi_{ t } &= 0 \notag
\end{align}
We seek for our regularization term to be scaled relative to the loss, so we perform a simple re-scaling:
\begin{align}\label{eq:regularization-rescale}
\regularizationTerm' &= \sqrt{ \max( \regularizationTerm, 1) ^ {\min(\max(s, 0), 1)} }, \\
s &= \frac{\log_e {\gL(\mW_{\rm{pos}}, \theta_{\rm{pos}}; \trainSplit)}}{\log_e{\regularizationTerm}} \notag
\end{align}
After rescaling, $\regularizationTerm'$ is an additive term in the next-token training loss, described below. Next, we describe the $\sdm$ activations, and the structure of the network, more generally.

\subsubsection{SDM network}\label{sec:sdm-network-details}

The network makes use of two separate $\sdm$ activations. The first ($\sdmVerificationLayer$) is over the binary verification task, trained at the document level. This is built as described in \S~\ref{sec:sdm-activation}, but specifically with an exemplar adaptor $g: \left ({\rm{mean}}(\vh_{\rm{neg}}), {\rm{mean}}(\vh_{\rm{pos}}), \vh^{-1}_{\rm{neg}}, \vh^{-1}_{\rm{pos}} \right ) \in \reals^{4D} \mapsto \vh' \in \reals^M$, trained over the concatenation of the mean of the final hidden states across tokens of both $\gM^{\rm{\rm{neg}}}$ and $\gM^{\rm{pos}}$, as well as the hidden state (i.e., $\vh^{-1}_{\rm{neg}} \in \reals^{D}$ and $\vh^{-1}_{\rm{pos}} \in \reals^{D}$) that predicts the end of sequence delimiter\footnote{This is the symbol that indicates the end of the sequence at the unit of analysis of the verification labels (e.g., at the sentence or document level).}, for which we use the superscript -1, all of which remain fixed when training the adaptor.\footnote{In our small-scale experiments, we only train the final hidden layer of the LLM (i.e., $\theta_{\rm{pos}}$ stays fixed, and we only update $\mW_{\rm{pos}}$), so we exclude the weights of $\gM^{\rm{neg}}$ as input to the exemplar adaptor, since they are identical to those of $\gM^{\rm{pos}}$. } This has an associated $\sdm$ estimator, $\EstimatorPLower$, over the binary verification task.

The second $\sdm$ activation is for normalizing the linear layer over the output vocabulary for next-token training, as described in \S~\ref{sec:negative+positive}. In this case, the output $\Magnitude$ is determined by the concatenation of $(\vz_{\rm{neg}}, \vz_{\rm{pos}})$, but the values of $\q$ and $d$ are from the $\sdmVerificationLayer$. In other words, for this second $\sdm$ activation, there is no exemplar adaptor inserted between the final hidden state of the LLM and the linear-layer over the vocabulary. This enables easily adapting this mechanism to existing architectures and pre-trained weights. 

\paragraph{SDM Network Next-token Loss.} Holding the weights of the $\sdmVerificationLayer$ fixed, the next token loss to update the weights of $\gM^{\rm{pos}}$, $\mW_{\rm{pos}}$ and $\theta_{\rm{pos}}$, is then:
\begin{align}\label{eq:sdm-network-next-token-training-loss}
\gL(\mW_{\rm{pos}}, \theta_{\rm{pos}}; \trainSplit, \beta, \gM^{\rm{ref}}) = -\frac{1}{N} \sum_{n}^{N} \log_{(2+q)}\left(
\frac{
(2+q)^{\dVal \cdot z_{{\rm{neg, pos}}_{t_n}}}
}{
\sum^{| \gV | \cdot 2}_{v=1}{(2+q)^{\dVal \cdot z_{{\rm{neg, pos}}_{v}}}}
}
\right)
+ \beta \regularizationTerm'
\end{align}
where $t_n$ is the index of the correct next token, and $\beta \in [0, \infty) $ linearly increases every mini-batch in an epoch from $\beta_{\rm{min}}$ (e.g., 0, in our experiments) to $\beta_{\rm{max}}$ (e.g., 0.1, in our experiments).

\paragraph{Train-time Generation vs. Train-time Force-decoding.} The loss in Eq.~\ref{eq:sdm-network-next-token-training-loss} requires $\q$ and $d$, which are predicated on labels at the document level, for each token prior to the model seeing the end of the document. In practice, for each $(\vx, y=1) \in \trainSplit$, prior to calculating the loss, we decode a completion for $\vx$ starting at the completion marker $x_t$ (e.g., starting at the instruction prompt, or given prefix, as noted in \S~\ref{sec:verification-encoding}) with $\q=e-2, d=1$. Then we derive $\q$ and $d$ from the $\sdmVerificationLayer$ over this generated output. We otherwise discard the generated completion and calculate the loss using these updated values of $\q$ and $d$ over the correct next token. (In the present work, $\q$ and $d$ are the same for each token in a single document.) Note that the stored support set of the $\sdmVerificationLayer$ (which determines $\q$ and $d$) is constructed by force-decoding over $(\vx, y=\{0, 1\}) \in \trainSplit$. Thus, the loss has the desired semantics of rewarding the model to resemble the $y=1$ data at the token-level (as in standard next-token fine-tuning), while penalizing generations that are challenging to verify. 

\paragraph{SDM Network Training: $\sdmVerificationLayer$ + Next-token Loop.}

The next-token loss and the $\sdmVerificationLayer$ interact via $\q$ and $d$ and the stopping criteria. However, the weight updates of each occur separately. 

We seek the weights that maximize the admitted points over $\calibrationSplit$ via $\EstimatorPLower$ for $\hat{y}=1$, and (if available), further restricting this set to those with correct $\yTask$ predictions (parsed from the generated text) for the underlying tasks encoded in the data.

The combined training loop is conceptually straightforward (Alg.~\ref{alg:sdm-network-training}). First, we construct the $\sdm$ estimator for binary verification ($\sdmVerificationLayer$) via Alg.~\ref{alg:sdm-estimator-training} by force-decoding over $\trainSplit$ and $\calibrationSplit$. (The convention is to shuffle $\trainSplit$ and $\calibrationSplit$ in the first training of the $\sdmVerificationLayer$, itself a process over $J$ iterations, and then use that final data split for all subsequent processes.) Next, we train one epoch of $\gM^{\rm{pos}}$. The next-token loss (Eq.~\ref{eq:sdm-network-next-token-training-loss}) uses $\q$ and $d$ from the $\sdmVerificationLayer$ over completions generated via greedy decoding (with $\q=e-2, d=1$) using $\sdm(\vz_{\rm{neg}}, \vz_{\rm{pos}})$ starting at the completion marker.\footnote{In our experiments, we cache $\q$ and $d$ before each epoch, but in principle, they can be calculated dynamically during an epoch as the weights change. Caching simplifies the implementation at the expense of potentially biasing the estimates as an epoch proceeds, which is the motivation for increasing $\beta$ through the course of an epoch.} Once the epoch concludes, we retrain the $\sdmVerificationLayer$ and update $\q$ and $d$ for $\trainSplit$. We then generate completions using $\sdm(\vz_{\rm{neg}}, \vz_{\rm{pos}})$ over $\calibrationSplit$ and calculate the number of points for which $\EstimatorPLower$ provides an index-conditional estimate for $\hat{y}=1$, further restricted (if applicable) to the underlying task labels, $\yTask$, and the predictions parsed for those tasks from the generated output. Next, we continue to the next epoch of updating $\gM^{\rm{pos}}$. This process continues until the max number of epochs has been reached.

\begin{algorithm}
    \caption{$\sdm$ Network Training}
    \label{alg:sdm-network-training}
    \small
    \begin{algorithmic}[1] 
        \Require $\trainSplit$, $\calibrationSplit$, $\alpha'$, $\text{max epochs}$, $\gM^{\rm{ref}}$, $\gM^{\rm{neg}}$, $\gM^{\rm{pos}}$ %
        \Procedure{sdm-network-train}{$\trainSplit$, $\calibrationSplit$, $\alpha'$, $\text{max epochs}$, $\gM^{\rm{ref}}$, $\gM^{\rm{neg}}$, $\gM^{\rm{pos}}$}
        		\State $\sdmVerificationLayer, \trainSplit_{*}, \calibrationSplit_{*}, \gE \gets$ \Call{sdm-iterative-train}{$\cdot$}  \Comment{ Alg.~\ref{alg:sdm-estimator-training}}
        		\State $\gM_{*} \gets$ Initialized with $\gM^{\rm{neg}}$, $\gM^{\rm{pos}}$ \Comment{Final trained model}
        		\State ${\rm{metric}}_{*} \gets 0$ \Comment{Determines final model}
		\State $\sdmVerificationLayer_{*} \gets \sdmVerificationLayer$ \Comment{Final $\sdm$ activation layer for verification}
        		\State $\gE_{*} \gets \gE$  \Comment{Final $\sdm$ estimator (i.e., $\EstimatorPLower$) for verification}
		\State $\beta_{\rm{step}} \gets \frac{\beta_{\rm{max}} - \beta_{\rm{min}}}{\rm{total~mini~batches}}$ \Comment{Used to calculate $\beta$ as a function of epoch progress}
		\State Calculate $q,d$ for each $(\vx, y=1) \in \trainSplit_{*}$ using $\sdmVerificationLayer$ over generated output from $\sdm(\vz_{\rm{neg}}, \vz_{\rm{pos}})$ with $\q=e-2, d=1$ \label{line:sdm-net-train-calculate-q-d}
		 \For{$e \in 1, \ldots, \text{max epochs}$}
		 	\State Minimize $\gL(\mW_{\rm{pos}}, \theta_{\rm{pos}}; \trainSplit, \beta, \gM^{\rm{ref}})$ \Comment{Eq.~\ref{eq:sdm-network-next-token-training-loss}}
			\State $\sdmVerificationLayer, \_, \_, \gE \gets$ \Call{sdm-iterative-train}{$\cdot$} \label{line:sdm-net-train-inner-sdm-retrain} \Comment{Without shuffling $\trainSplit_{*}, \calibrationSplit_{*}$}
			\State Update $q,d$ for each $(\vx, y=1) \in \trainSplit_{*}$ \Comment{As in Line~\ref{line:sdm-net-train-calculate-q-d}}
			\State $\rm{metric} \gets$ cardinality of the admitted set from $\EstimatorPLower$ for $\hat{y}=1$ over $\calibrationSplit_{*}$ \label{line:sdm-net-train-stopping-criteria} \Comment{Restricted to $\yTask=\yTaskPrediction$, if available}
          	 	\If{$\rm{metric} > \rm{metric}_{*}$}
           			\State $\rm{metric}_{*} \gets \rm{metric}$
				\State $\gM_{*} \gets$ Update with $\mW_{\rm{pos}}, \theta_{\rm{pos}}$
				\State $\sdmVerificationLayer_{*} \gets \sdmVerificationLayer$
				\State $\gE_{*} \gets \gE$
           		\EndIf
		 \EndFor
	\State \textbf{return} $\gM_{*}, \trainSplit_{*}, \calibrationSplit_{*}, \sdmVerificationLayer_{*}, \gE_{*}$
        \EndProcedure
        \Ensure $\gM_{*}, \trainSplit_{*}, \calibrationSplit_{*}, \sdmVerificationLayer_{*}, \gE_{*}$ 
    \end{algorithmic}
\end{algorithm}

\paragraph{SDM Network Test-time Generation.}

At test time, we generate from $\sdm(\vz_{\rm{neg}}, \vz_{\rm{pos}})$ up to the output control token, or end-of-sequence token, at the unit-of-analysis of the verification labels, via greedy (i.e., $\argmax$) decoding with $\q=e-2, d=1$ (i.e., equivalent to $\softmax$).\footnote{As previously noted, at test time, $\left ( \argmax(\sdm(\vz_{\rm{neg}}, \vz_{\rm{pos}}))~{\rm{mod}}~|\gV| \right)$ maps back to the original token symbol in the vocabulary when decoding over the joint distribution.} We then continue generation, or take other branching actions, based on $\EstimatorPLower$ from the $\sdmVerificationLayer$, which by extension, also provides interpretability-by-exemplar into $\trainSplit$ via matching (from $\q$) and against similarly calibrated  points in $\calibrationSplit$ via $\hardqbin$. Each classification via the $\sdmVerificationLayer$ requires on the order of the computation needed for commonly used dense retrieval augmentations of LLMs, so such test-time generation and verification is achievable even using edge devices.

\section{Experiments}
We comprehensively evaluate the uncertainty-awareness of our estimators across a representative set of the existing classes of estimators over LLMs. First, we compare $\sdm$ calibration to existing approaches in a standard classification setting, using open-source models at a scale that can be readily replicated with consumer-level compute (\S~\ref{sec:experiments-classification}). Next, we show how an $\sdm$ estimator can be applied to a fully black-box LLM API, only with access to the top output logits and without a proxy model running in parallel, using the standard $\datasetMMLU$ benchmark (\S~\ref{sec:experiments-black-box}). In this context, we also consider a data quality experiment in which we seek to detect errors in the carefully curated \textsc{MMLU-Pro} dataset. This serves as a natural, held-out blind evaluation of the estimator's capacity to separate aleatoric and epistemic uncertainty. Finally, we examine the universal verification behavior of an $\sdm$ network by training over a composition of the classification tasks examined in the first set of targeted experiments (\S~\ref{ref:experiments-universal-verification}).

\subsection{Experiments: Classification}\label{sec:experiments-classification}

Before introducing the additional complications of LLM generation, we first isolate the core calibration behavior against existing classes of approaches in standard multi-class classification settings.

\subsubsection{Task: $\datasetSentiment$} 
\paragraph{Task.}
Our first task ($\datasetSentiment$) is predicting the sentiment of movie reviews using the commonly used benchmark data of \citet{Maas-EtAl-2011-OriginalCitedSourceForIMDbReviewsData}. This is a binary classification task with $y \in \{0 = {\rm{negative}}, 1 = {\rm{positive}}\}$. $\trainSplit$ and $\calibrationSplit$ are constructed from a total of 18k instances. The held-out set for evaluation, $|\testSplit|=1583$, is from the same distribution as $\trainSplit$ and $\calibrationSplit$. This is a well-studied task for which the surface-level signals correlated with the target labels are expected to be effectively modeled by large parameter LLMs; as such, relatively high task accuracies are expected.

\paragraph{Models.} Our base $\underlyingNetwork$, the parameters of which stay fixed and are used for all estimators, is the open-source, publicly available \texttt{Faster I} model from the on-device data analysis program \textsc{Reexpress one}
 from Reexpress AI. This 1.2 billion-parameter model is a late fusion of the encoder and decoder of Flan-T5 large \citep{flanT5-2022} and mT0-base \citep{muennighoff-etal-2023-crosslingual}. We discard the existing adaptor layers that are part of the on-device program and only use the parameter fusion of the encoder and decoder, adding the adaptors and estimators introduced in this work. We take the mean of the hidden states across input tokens, resulting in a hidden state of $\vh \in \reals^{3774}$ as input to either an exemplar adaptor, or an $\sdm$ activation layer, each with $M=1000$. We use the label $\modelFasterICNNAdaptor$ for a standard exemplar adaptor over $\vh \in \reals^{3774}$ trained with a cross-entropy loss, and the label $\modelFasterISDM$ for the $\sdm$ activation layer over $\vh \in \reals^{3774}$.

\paragraph{Estimators.} Holding the underlying $\underlyingNetwork$ constant, we examine representative classes of estimators used with neural networks, seeking index-conditional calibration at $\alpha'=0.95$. At the most basic, but also, perhaps the most commonly used in practice, representing the absence of a post-hoc calibration method, we simply threshold the output, $\softmax(\vz) \ge \alpha'$, where the temperature $\tau=1$. As an established empirical approach for calibrating neural networks, we provide a comparison to temperature scaling \citep{GuoEtAl-2017-TempScaling}, a single parameter version of post-hoc Platt-scaling \citep{Platt-1999-PlattScaling}, with the label $\estimatorTempScaling$. In this case, the estimator is the thresholding of the output $\softmax(\vz; \tau) \ge \alpha'$ after learning a value for $\tau$ over $\calibrationSplit$. We also provide a comparison to two representative conformal predictors, the $\conformalAPS$ method of \citet{RomanoEtAl-2020-APS} and the adaptiveness-optimized $\conformalRAPS$ algorithm of \citet{Angelopoulos-2021-RAPS}. The admission criteria for the $\conformalAPS$ and $\conformalRAPS$ estimators is prediction sets of size 1, using an $\alpha=0.05$. 

We then compare to the primary $\sdm$ estimator $\EstimatorPLower$, as well as the reference comparisons $\EstimatorPCentroid$ and $\EstimatorPUpper$, as defined in \S~\ref{sec:test-time-estimates}. We train the $\sdm$ activation layer and estimator (Alg.~\ref{alg:sdm-estimator-training}) with $J=10$, here and for the remaining experiments. Additional training hyper-parameters and details shared across all experiments are provided in Appendix~\ref{appendix:additional-training-details}.

As a common point of reference, here and for all other experiments as well, we will use the label $\estimatorNoReject$ to refer to the model predictions without any selective filtering (i.e., the raw output accuracies, either from a $\softmax$ or an $\sdm$ activation). 
  
\subsubsection{Task: $\datasetSentimentOOD$} 

To evaluate the behavior of the estimators over out-of-distribution data, we consider an additional task ($\datasetSentimentOOD$) that uses the same models and estimators as $\datasetSentiment$, but an out-of-distribution evaluation set, $|\testSplit|=4750$. We use the SemEval-2017 Task 4a test set \citep{Rosenthal-etal-2017-Semeval-Task4}, which consists of short-form social media posts that differ in the distribution of topics, language styles, and lengths relative to the movie reviews. We balance the test set, dropping the third class (neutral), setting the semantics of the true labels to be the same as that of the movie reviews: $y \in \{0 = {\rm{negative}}, 1 = {\rm{positive}}\}$.
 
\subsubsection{Task: $\datasetFactcheck$} 

\paragraph{Task.}
As a more challenging binary classification task for LLMs, we consider the fact check data of \citet{azaria-mitchell-2023-internal}. The training and calibration sets, a combined total of 6k instances, consist of single sentence statements that have been semi-automatically generated via templates and a knowledge base. The task is to determine whether the statement is true or false,  $y \in \{0 = {\rm{false}}, 1 = {\rm{true}}\}$. The held-out eval set, $|\testSplit|=245$, the focus of our analysis, has been constructed by having an LLM generate a statement continued from a true statement not otherwise in the dataset. These evaluation statements are checked manually and assigned labels by human annotators. In addition to being a relatively challenging task that evaluates---at least in principle---the latent knowledge stored within an LLM's parameters, the test set is representative of the types of distribution shifts over high-dimensional inputs that can be problematic for real applications, and challenging to characterize without model assistance and ground-truth labels. It was observed in \citet{azaria-mitchell-2023-internal} that the accuracy of existing LLM classifiers is dramatically lower on this generated, held-out test set compared to the calibration set. However, these test sentences would seem to also be simple true-false statements, reflecting that it is not always immediately obvious for a human user to detect distribution shifts over high-dimensional inputs. As such, we seek for our models and estimators to reflect such shifts via the predictive uncertainty, as we will not, in general, have true labels at test time.

\paragraph{Models and Estimators.} Reflecting the more challenging task, our base $\underlyingNetwork$ is the larger 3.2 billion parameter \texttt{Fast I} model from \textsc{Reexpress one}, which is a late fusion of the encoder and decoder of Flan-T5 xl and mT0-base. We additionally compose the \texttt{Fast I} model with \texttt{Mixtral 8x7B Instruct v0.1} \citep{JiangEtAl-2024-Mixtral8x7B}. This is achieved by constructing a simple re-ask verification prompt, and then a transform of the final layer of the \texttt{Mixtral} model and the output logits is concatenated to the mean of the hidden states across the input tokens of \texttt{Fast I}. 
We use the label $\modelFastIMixtralCNNAdaptor$ for a standard exemplar adaptor over the resulting $\vh \in \reals^{5854}$ trained with a cross-entropy loss, and the label $\modelFastIMixtralSDM$ for the $\sdm$ activation layer over $\vh \in \reals^{5854}$. The estimators are otherwise the same as those used for the $\datasetSentiment$ task.

\subsection{Experiments: Black-box LLM APIs}\label{sec:experiments-black-box}

Next, we examine the behavior of the estimators when we only have access to a black-box API for an LLM that provides the generated text and the top-1 output log probabilities. In this context with a state-of-the-art model, we examine an additional class of estimators: Those that make use of uncertainty estimates explicitly encoded in the surface-level output vocabulary symbols. As a fully held-out test---and real-world use example---we also consider a data quality experiment in which we seek to uncover annotation errors in an existing carefully curated benchmark dataset.

\subsubsection{Task: Question Answering}

\paragraph{Task.} Our evaluation is over the 4-choice question answering benchmark dataset $\datasetMMLU$ \citep{HendrycksEtAl-2021-MMLU} and a 4-choice subset of the more challenging \textsc{MMLU-Pro} dataset \citep{WangEtAl-2024-MMLUpro}\footnote{Both datasets are available via \url{https://huggingface.co/datasets}}, for which we use the label $\datasetMMLUPro4qa$. $\trainSplit$ and $\calibrationSplit$ are constructed from 102k instances from the \texttt{auxiliary\_train}, \texttt{dev}, and \texttt{val} splits of $\datasetMMLU$ and the MMLU-Pro validation set, the 4-choice subset of which only consists of 29 instances. For $\datasetMMLU$, $|\testSplit|=14042$. For $\datasetMMLUPro4qa$, $|\testSplit|=5413$.

\paragraph{Models and Estimators.} We use \texttt{gpt-4o-2024-08-06} (\textsc{gpt-4o}) \citep{openai-2024-gpt4ocard} via the Microsoft Azure service\footnote{\url{https://azure.microsoft.com/en-us/}} as the black-box LLM. Given the zero-shot question, the LLM is tasked with providing a structured response against the JSON Schema in Listing~\ref{lst:llm-json-schema}, and the top-1 log probability for each output token. The JSON is parsed for the answer letter, the surface-level symbol of which is the prediction for the $\estimatorNoReject$ estimator of $\modelGPTFourO$. We consider the output probability for the answer letter, restricted to those estimates $\ge \alpha'$, as $\estimatorLetterProb$. The output JSON is also parsed for the model's real-valued verbalized uncertainty estimate, which when restricted to estimates $\ge \alpha'$, is the estimator $\estimatorVerbProb$.\footnote{An additional class of estimators for black-box LLMs are those that require multiple test-time forward passes through the model, which are related to the Bayesian approaches of \citet[inter alia]{GalAndZoubin-2016-MCDropout}. We do not consider this class of approaches given the computational costs required of the estimators.}

As a final field, the output JSON also contains a short explanation for the response. We take the mean of the output probabilities corresponding to each value of the output JSON and concatenate those three values with a soft feature vector of length 4, where the activated index is that of the surface-level answer choice, for which we use $\estimatorVerbProb$ as the value, and all other indexes are 0. This length 7 vector than serves as $\vh \in \reals^{7}$ as input to an $\sdm$ activation layer with $M=1000$. For the resulting $\modelGPTFourOSDM$ model, we consider the $\estimatorNoReject$, $\EstimatorPLower$, $\EstimatorPCentroid$, and $\EstimatorPUpper$ estimators. Additional details appear in Appendix~\ref{Appendix:LLM-APIs}.

\subsubsection{Task: Data Quality Analysis}

The MMLU-Pro dataset ($\datasetMMLUPro4qa$) is a follow-up to the original $\datasetMMLU$ benchmark designed to have more challenging questions and more reliable answer annotations. In the previously described experiment, we examine whether calibration can be maintained over this implied distribution shift. Separately, we consider here whether our method can uncover additional annotation errors, despite the relatively large amount of resources already spent to refine the dataset by the dataset constructors. MMLU-Pro reportedly underwent multiple rounds of review with experts and annotators, including LLM assistance for targeted error detection. We focus on the Computer Science category given that the questions should have unambiguous, objectively verifiable answers. This data quality test is a natural, fully held-out assessment of our approach compared to existing approaches used in practice, with direct, real-world applications. To do so we will examine the annotations among the set of admitted points sorted by $\EstimatorPLower$ for which $y \ne \hat{y}$, where the desired behavior is for these points to reflect the aleatoric uncertainty (exogenous to the model and estimator) of label annotation errors.

\subsection{Experiments: Verified Generation}\label{ref:experiments-universal-verification}

Next, given the context of the above experiments, we examine the behavior of the $\sdm$ network.

\paragraph{Task.} We construct the verification task from the $\datasetSentiment$, $\datasetSentimentOOD$, and $\datasetFactcheck$ data described above (\S~\ref{sec:experiments-classification}), taking the $y$ labels of those earlier tasks as the $\yTask$ labels. The $\yVerification$ (or simply $y$) labels (and associated instances) are constructed by synthetically inverting the text of the associated completions, as illustrated in Table~\ref{tab:experiments-verified-generation-data-format}. By design, under the assumption that it is a more challenging learning setting, we do not pair the completions. For example, given a single movie review, it will appear once as part of a user prompt and either the label $\yVerification=0$ or $\yVerification=1$, but not both.\footnote{
As noted in \S~\ref{sec:sdm-networks}, $y=0$ instances (here, the constructed negatives) do not directly participate in next-token fine-tuning, but they are used for training the $\sdmVerificationLayer$ and the support set to determine $\q$ and $d$.
} 

For analysis, we then have a standard binary classification task over the force-decoded output, $(\vx, \yVerification \in \{0,1\}) \in \testSplit$. We use the following labels for the corresponding datasets: $\datasetSentimentVerification$, with $|\testSplit|=1583$; $\datasetSentimentOODVerification$, with $|\testSplit|=4750$; and $\datasetFactcheckVerification$, with $|\testSplit|=245$. These test sets are useful for analyzing the behavior of the $\sdmVerificationLayer$, but they do not reflect a real test-time scenario.

For final evaluation, we take the original test sets from $\datasetSentiment$, $\datasetSentimentOOD$, and $\datasetFactcheck$ (\S~\ref{sec:experiments-classification}) and evaluate the output of the generated JSON for the underlying task labels, $\yTask$, as in a standard evaluation of LLM output.

The corresponding system and user prompts appear in Listing~\ref{lst:verified-generation-inputs}. These design decisions enable examining the instruction-following setting across multiple underlying tasks while enabling reliable evaluation of verification, since there is no ambiguity (up to annotation errors in the original tasks) in $\yTask$ and $\yVerification$, and we can readily parse the JSON output for the task predictions.\footnote{Ill-formatted JSON output is treated as a wrong prediction.}

\paragraph{Models and Estimators.} For $\gM^{\rm{ref}}$ we use \texttt{Phi-3.5-mini-instruct} model ($\modelPhiThreeFiveInstruct$) \citep{Abdin-2024-Phi3-TechReport}, a 3.8 billion-parameter decoder-only Transformer model, via MLX \citep{HannunEtAl-MLX-2023}, version 0.21.1. To keep the experiments manageable at a level of compute that can be readily replicated on consumer hardware, while still being instructive for future larger-scale experiments, we only update the final linear-layer of $\gM^{\rm{pos}}$, $\mW_{\rm{pos}}$, in the next-token loss (Eq.~\ref{eq:sdm-network-next-token-training-loss}); however, we update the full weight matrix of $\mW_{\rm{pos}}$ and not a lower-rank adaptor over these weights. This is instructive in this context, since our data is relatively small, but with $| \gV | = 32064$ and the $\modelPhiThreeFiveInstruct$ hidden dimension of $3072$, the 100 million parameters of $\mW_{\rm{pos}}$ would be assumed to quickly overfit, leading to degenerate output. Because we only update $\mW_{\rm{pos}}$, while $\theta_{\rm{pos}}$ stays fixed, we only need to train the $\sdmVerificationLayer$ once before the next-token training loop begins (i.e., Line~\ref{line:sdm-net-train-inner-sdm-retrain} in Alg.~\ref{alg:sdm-network-training} is not needed), and we exclude the weights of $\gM^{\rm{neg}}$ as input to the $\sdm$ activation layer, since they are identical to those of $\gM^{\rm{pos}}$. As such, the input to the $\sdm$ activation of the $\sdmVerificationLayer$ is $\left ({\rm{mean}}(\vh_{\rm{pos}}), \vh^{-1}_{\rm{pos}} \right ) \in \reals^{2\cdot 3072}$, the concatenation of the average of the final hidden states (across tokens) with the final hidden state that predicts the end of sequence delimiter (here, the final closing bracket in the JSON output).

In this setting, our primary comparison is against the full model before fine-tuning, for which we use the label $\modelPhiThreeFiveInstructSDM$. In this case, only the $\sdmVerificationLayer$ layer is trained, here for $J=10$ iterations of 50 epochs, but the evaluation is still over completions generated via greedy decoding (i.e., $\argmax$) over $\sdm(\vz_{\rm{neg}}, \vz_{\rm{pos}})$ with $\q=e-2, d=1$. The fine-tuned model ($\modelPhiThreeFiveInstructSDMNetwork$) uses this same $\sdmVerificationLayer$, but it is also trained for 5 epochs with $\beta_{\rm{min}}=0$ and $\beta_{\rm{max}} = 0.1$ using the next-token loss of Alg.~\ref{alg:sdm-network-training}. We choose the model weights (as in Line~\ref{line:sdm-net-train-stopping-criteria} of Alg.~\ref{alg:sdm-network-training}) as those that maximize the count of admitted points over $\calibrationSplit$ via $\EstimatorPLower$ for $\hat{y}=1$, further restricted to $\yTask=\yTaskPrediction$, which is determined by parsing the generated JSON output. For both models, $\modelPhiThreeFiveInstructSDM$ and $\modelPhiThreeFiveInstructSDMNetwork$, we consider the $\estimatorNoReject$ and $\EstimatorPLower$ estimators.\footnote{
In this context, $\EstimatorPCentroid$ and $\EstimatorPUpper$ are less meaningful as a comparison since $\EstimatorPLower$ is used as part of the aforementioned stopping criteria for $\modelPhiThreeFiveInstructSDMNetwork$, and are thus excluded.
}
 
\section{Results}
\begin{table}
  \caption{Comparison of relevant estimators for the standard document classification setting, $\alpha'=\colorbox{correctPredictionColor}{0.95}$. \colorbox{correctPredictionColor}{N/A} indicates all predictions were rejected, which is preferred over falling under the expected accuracy. 
  $n=|\text{Admitted}|$, the count of non-rejected documents.} %
  \label{tab:experiments-classification-combined} 
  \centering
  \resizebox{1.0\textwidth}{!}{%
  \begin{tabular}{l l  l c c c c c c c c c l }
    \toprule

    & & &  \multicolumn{4}{c}{Class-conditional} & \multicolumn{4}{c}{Prediction-conditional}  & \multicolumn{2}{c}{Marginal} \\
    & & &  \multicolumn{2}{c}{$y=0$} & \multicolumn{2}{c}{$y=1$} & \multicolumn{2}{c}{$\hat{y}=0$} & \multicolumn{2}{c}{$\hat{y}=1$} & \multicolumn{2}{c}{$y\in \{0,1\}$} \\
    \cmidrule(r){4-5} \cmidrule(r){6-7} \cmidrule(r){8-9} \cmidrule(r){10-11} \cmidrule(r){12-13} \\
    Dataset   & Model & Estimator & \textsc{Acc.}& $\frac{n}{|\testSplit|}$ & \textsc{Acc.} & $\frac{n}{|\testSplit|}$ & \textsc{Acc.} & $\frac{n}{|\testSplit|}$ & \textsc{Acc.} & $\frac{n}{|\testSplit|}$ & \textsc{Acc.} & $\frac{n}{|\testSplit|}$\\
    \midrule
    $\datasetSentiment$ & $\modelFasterICNNAdaptor$ & $\estimatorNoReject$ & \colorbox{correctPredictionColor}{0.982} & 0.50 & \colorbox{correctPredictionColor}{0.953} & 0.50 & \colorbox{correctPredictionColor}{0.955} & 0.51 & \colorbox{correctPredictionColor}{0.982} & 0.49 & \colorbox{correctPredictionColor}{0.968} & 1. \\
    $\datasetSentiment$ & $\modelFasterICNNAdaptor$ & $\estimatorSoftmax$ & \colorbox{correctPredictionColor}{0.995} & 0.46 & \colorbox{correctPredictionColor}{0.983} & 0.41 & \colorbox{correctPredictionColor}{0.985} & 0.46 & \colorbox{correctPredictionColor}{0.994} & 0.41 & \colorbox{correctPredictionColor}{0.989} & 0.87 \\
    
    $\datasetSentiment$ & $\modelFasterICNNAdaptor$ & $\estimatorTempScaling$ & \colorbox{correctPredictionColor}{0.994} & 0.45 & \colorbox{correctPredictionColor}{0.986} & 0.39 & \colorbox{correctPredictionColor}{0.987} & 0.45 & \colorbox{correctPredictionColor}{0.994} & 0.39 & \colorbox{correctPredictionColor}{0.990} & 0.84 \\
    
    $\datasetSentiment$ & $\modelFasterICNNAdaptor$ & $\conformalAPS$ & \colorbox{correctPredictionColor}{0.993} & 0.47 & \colorbox{correctPredictionColor}{0.973} & 0.45 & \colorbox{correctPredictionColor}{0.975} & 0.48 & \colorbox{correctPredictionColor}{0.993} & 0.44 & \colorbox{correctPredictionColor}{0.983} & 0.92 \\
    
    $\datasetSentiment$ & $\modelFasterICNNAdaptor$ & $\conformalRAPS$ & \colorbox{correctPredictionColor}{0.989} & 0.47 & \colorbox{correctPredictionColor}{0.972} & 0.44 & \colorbox{correctPredictionColor}{0.974} & 0.48 & \colorbox{correctPredictionColor}{0.988} & 0.44 & \colorbox{correctPredictionColor}{0.981} & 0.92 \\
    
    $\datasetSentiment$ & $\modelFasterISDM$ & $\estimatorNoReject$ & \colorbox{correctPredictionColor}{0.971} & 0.50 & \colorbox{correctPredictionColor}{0.966} & 0.50 & \colorbox{correctPredictionColor}{0.966} & 0.50 & \colorbox{correctPredictionColor}{0.971} & 0.50 & \colorbox{correctPredictionColor}{0.968} & 1. \\
    
    $\datasetSentiment$ & $\modelFasterISDM$ & $\EstimatorPLower$ & \colorbox{correctPredictionColor}{0.996} & 0.32 & \colorbox{correctPredictionColor}{0.996} & 0.32 & \colorbox{correctPredictionColor}{0.996} & 0.32 & \colorbox{correctPredictionColor}{0.996} & 0.32 & \colorbox{correctPredictionColor}{0.996} & 0.65 \\
    
    $\datasetSentiment$ & $\modelFasterISDM$ & $\EstimatorPCentroid$ & \colorbox{correctPredictionColor}{0.996} & 0.36 & \colorbox{correctPredictionColor}{0.993} & 0.35 & \colorbox{correctPredictionColor}{0.993} & 0.36 & \colorbox{correctPredictionColor}{0.996} & 0.35 & \colorbox{correctPredictionColor}{0.995} & 0.71 \\
    
    $\datasetSentiment$ & $\modelFasterISDM$ & $\EstimatorPUpper$ & \colorbox{correctPredictionColor}{0.997} & 0.38 & \colorbox{correctPredictionColor}{0.993} & 0.37 & \colorbox{correctPredictionColor}{0.993} & 0.38 & \colorbox{correctPredictionColor}{0.997} & 0.37 & \colorbox{correctPredictionColor}{0.995} & 0.75 \\
  \midrule 
  $\datasetSentimentOOD$ & $\modelFasterICNNAdaptor$ & $\estimatorNoReject$ & \colorbox{correctPredictionColor}{0.992} & 0.5~~ & \colorbox{wrongPredictionColor}{0.394} & 0.5~~ & \colorbox{wrongPredictionColor}{0.621} & 0.80 & \colorbox{correctPredictionColor}{0.979} & 0.20 & \colorbox{wrongPredictionColor}{0.693} & 1. \\
  $\datasetSentimentOOD$ & $\modelFasterICNNAdaptor$ & $\estimatorSoftmax$ & \colorbox{correctPredictionColor}{1.~~~~~~} & 0.37 & \colorbox{wrongPredictionColor}{0.251} & 0.08 & \colorbox{wrongPredictionColor}{0.854} & 0.44 & \colorbox{correctPredictionColor}{1.~~~~~~} & 0.02 & \colorbox{wrongPredictionColor}{0.861} & 0.46 \\
  
  $\datasetSentimentOOD$ & $\modelFasterICNNAdaptor$ & $\estimatorTempScaling$ & \colorbox{correctPredictionColor}{1.~~~~~~} & 0.34 & \colorbox{wrongPredictionColor}{0.223} & 0.07 & \colorbox{wrongPredictionColor}{0.869} & 0.39 & \colorbox{correctPredictionColor}{1.~~~~~~} & 0.01 & \colorbox{wrongPredictionColor}{0.874} & 0.41 \\
  
  $\datasetSentimentOOD$ & $\modelFasterICNNAdaptor$ & $\conformalAPS$ & \colorbox{correctPredictionColor}{1.000} & 0.43 & \colorbox{wrongPredictionColor}{0.346} & 0.19 & \colorbox{wrongPredictionColor}{0.770} & 0.55 & \colorbox{correctPredictionColor}{0.997} & 0.07 & \colorbox{wrongPredictionColor}{0.795} & 0.62 \\
  
  $\datasetSentimentOOD$ & $\modelFasterICNNAdaptor$ & $\conformalRAPS$ & \colorbox{correctPredictionColor}{0.999} & 0.43 & \colorbox{wrongPredictionColor}{0.336} & 0.20 & \colorbox{wrongPredictionColor}{0.761} & 0.56 & \colorbox{correctPredictionColor}{0.991} & 0.07 & \colorbox{wrongPredictionColor}{0.786} & 0.63 \\
  
  $\datasetSentimentOOD$ & $\modelFasterISDM$ & $\estimatorNoReject$ & \colorbox{wrongPredictionColor}{0.570} & 0.5~~ & \colorbox{correctPredictionColor}{0.966} & 0.5~~ & \colorbox{correctPredictionColor}{0.944} & 0.30 & \colorbox{wrongPredictionColor}{0.692} & 0.70 & \colorbox{wrongPredictionColor}{0.768} & 1. \\
  
  $\datasetSentimentOOD$ & $\modelFasterISDM$ & $\EstimatorPLower$ & \colorbox{correctPredictionColor}{~N/A~} & 0.~~~~ & \colorbox{correctPredictionColor}{~N/A~} & 0.~~~~ & \colorbox{correctPredictionColor}{~N/A~} & 0.~~~~ & \colorbox{correctPredictionColor}{~N/A~} & 0.~~~~ & \colorbox{correctPredictionColor}{~N/A~} & 0. \\
  $\datasetSentimentOOD$ & $\modelFasterISDM$ & $\EstimatorPCentroid$ & \colorbox{correctPredictionColor}{~N/A~} & 0.~~~~ & \colorbox{correctPredictionColor}{~N/A~} & 0.~~~~ & \colorbox{correctPredictionColor}{~N/A~} & 0.~~~~ & \colorbox{correctPredictionColor}{~N/A~} & 0.~~~~ & \colorbox{correctPredictionColor}{~N/A~} & 0. \\
  $\datasetSentimentOOD$ & $\modelFasterISDM$ & $\EstimatorPUpper$ & \colorbox{wrongPredictionColor}{0.~~~~~~} & 0.06 & \colorbox{correctPredictionColor}{1.~~~~~~} & 0.28 & \colorbox{correctPredictionColor}{~N/A~} & 0.~~~~ & \colorbox{wrongPredictionColor}{0.819} & 0.35 & \colorbox{wrongPredictionColor}{0.819} & 0.35 \\
  \midrule 
 $\datasetFactcheck$ & $\modelFastIMixtralCNNAdaptor$ & $\estimatorNoReject$ & \colorbox{wrongPredictionColor}{0.365} & 0.51 & \colorbox{wrongPredictionColor}{0.908} & 0.49 & \colorbox{wrongPredictionColor}{0.807} & 0.23 & \colorbox{wrongPredictionColor}{0.574} & 0.77 & \colorbox{wrongPredictionColor}{0.629} & 1. \\
 
 $\datasetFactcheck$ & $\modelFastIMixtralCNNAdaptor$ & $\estimatorSoftmax$ & \colorbox{wrongPredictionColor}{0.211} & 0.08 & \colorbox{correctPredictionColor}{0.975} & 0.33 & \colorbox{wrongPredictionColor}{0.667} & 0.02 & \colorbox{wrongPredictionColor}{0.839} & 0.38 & \colorbox{wrongPredictionColor}{0.828} & 0.40 \\
 
 $\datasetFactcheck$ & $\modelFastIMixtralCNNAdaptor$ & $\estimatorTempScaling$ & \colorbox{wrongPredictionColor}{0.286} & 0.06 & \colorbox{correctPredictionColor}{0.987} & 0.31 & \colorbox{wrongPredictionColor}{0.8~~~~} & 0.02 & \colorbox{wrongPredictionColor}{0.884} & 0.35 & \colorbox{wrongPredictionColor}{0.879} & 0.37 \\
 
 $\datasetFactcheck$ & $\modelFastIMixtralCNNAdaptor$ & $\conformalAPS$ & \colorbox{wrongPredictionColor}{0.283} & 0.19 & \colorbox{correctPredictionColor}{0.979} & 0.38 & \colorbox{wrongPredictionColor}{0.867} & 0.06 & \colorbox{wrongPredictionColor}{0.736} & 0.51 & \colorbox{wrongPredictionColor}{0.75~~} & 0.57 \\
 
 $\datasetFactcheck$ & $\modelFastIMixtralCNNAdaptor$ & $\conformalRAPS$ & \colorbox{wrongPredictionColor}{0.341} & 0.18 & \colorbox{correctPredictionColor}{0.967} & 0.37 & \colorbox{wrongPredictionColor}{0.833} & 0.07 & \colorbox{wrongPredictionColor}{0.75~~} & 0.47 & \colorbox{wrongPredictionColor}{0.761} & 0.55 \\

 $\datasetFactcheck$ & $\modelFastIMixtralSDM$ & $\estimatorNoReject$ & \colorbox{wrongPredictionColor}{0.397} & 0.51 & \colorbox{wrongPredictionColor}{0.899} & 0.49 & \colorbox{wrongPredictionColor}{0.806} & 0.25 & \colorbox{wrongPredictionColor}{0.585} & 0.75 & \colorbox{wrongPredictionColor}{0.641} & 1. \\
 
 $\datasetFactcheck$ & $\modelFastIMixtralSDM$ & $\EstimatorPLower$ & \colorbox{correctPredictionColor}{~N/A~} & 0.~~~~ & \colorbox{correctPredictionColor}{1.~~~~~~} & 0.13 & \colorbox{correctPredictionColor}{~N/A~} & 0.~~~~ & \colorbox{correctPredictionColor}{1.~~~~~~} & 0.13 & \colorbox{correctPredictionColor}{1.~~~~~~} & 0.13 \\
 
 $\datasetFactcheck$ & $\modelFastIMixtralSDM$ & $\EstimatorPCentroid$ & \colorbox{correctPredictionColor}{~N/A~} & 0.~~~~ & \colorbox{correctPredictionColor}{1.~~~~~~} & 0.17 & \colorbox{correctPredictionColor}{~N/A~} & 0.~~~~ & \colorbox{correctPredictionColor}{1.~~~~~~} & 0.17 & \colorbox{correctPredictionColor}{1.~~~~~~} & 0.17 \\
  
 $\datasetFactcheck$ & $\modelFastIMixtralSDM$ & $\EstimatorPUpper$ & \colorbox{correctPredictionColor}{1.~~~~~~} & 0.02 & \colorbox{correctPredictionColor}{0.980} & 0.21 & \colorbox{wrongPredictionColor}{0.8~~~~} & 0.02 & \colorbox{correctPredictionColor}{1.~~~~~~} & 0.20 & \colorbox{correctPredictionColor}{0.982} & 0.22 \\
   

    \bottomrule
  \end{tabular}
  }  
\end{table}
Across tasks and models, the $\sdm$ calibration process yields an estimator that achieves index-conditional calibration (Def.~\ref{defn:index-conditional-calibration}), in contrast to the existing classes of estimators over LLMs, which become unreliable in the presence of even modest distribution shifts. The $\EstimatorPLower$ estimator remains calibrated in the presence of distribution shifts due to the $\CauchyMinValidQBin$ lower constraint on $\softqbinLower$, which screens points that are unlike those seen during the calibration process. With existing methods, defining an out-of-distribution point has been task- and problem-specific, and generally challenging over high-dimensional inputs. In contrast, the $\sdm$ calibration process provides a principled approach for determining such cut-offs in a data- and model-driven manner, with minimal hyper-parameters, resulting in a clear separation of points over which the estimator is reliable (namely, the admitted points) and those over which the estimates themselves are unreliable (i.e., the rejected points). The $\sdm$ network incorporates this behavior into the LLM architecture and fine-tuning process to serve as a universal verifier, suggesting a principled basis for building large, complex LLM systems and pipelines that are reliable and interpretable with respect to the observed labeled data.

\subsection{Results: Classification}

\begin{table}
  \caption{$\CauchyMAD$ and $\CauchyMagnitudeIteratedLowerEstimate$ by $\hardqbin$ on $\calibrationSplit$ for the standard classification tasks, trained with $J=10$ iterations, each of 50 epochs. As $\hardqbin$ increases, the variation across instances decreases.} 
  \label{tab:experiments-classification-hard-qbin-iterated} 
  \centering
  \resizebox{0.7\textwidth}{!}{%
  \begin{tabular}{c  l l l l l l l l}
    \toprule
&  \multicolumn{4}{c}{$\datasetSentiment$} & \multicolumn{4}{c}{$\datasetFactcheck$} \\
     \cmidrule(r){2-5} \cmidrule(r){6-9}\\
     &  \multicolumn{2}{c}{$y=0$} & \multicolumn{2}{c}{$y=1$} & \multicolumn{2}{c}{$y=0$} & \multicolumn{2}{c}{$y=1$} \\
     \cmidrule(r){2-3} \cmidrule(r){4-5} \cmidrule(r){6-7} \cmidrule(r){8-9} \\
    $\hardqbin$    & $\CauchyMAD$ & $\CauchyMagnitudeIteratedLowerEstimate[0]$ & $\CauchyMAD$ & $\CauchyMagnitudeIteratedLowerEstimate[1]$ & $\CauchyMAD$ & $\CauchyMagnitudeIteratedLowerEstimate[0]$ & $\CauchyMAD$ & $\CauchyMagnitudeIteratedLowerEstimate[1]$\\
    \midrule
       0 & 0.007 & 0.044 & 0.007 & 0.044 & 0.024 & 0.148 & 0.018 & 0.116 \\
   1 & < 0.001 & 0.006 & < 0.001 & 0.003 & 0.009 & 0.056 & 0.004 & 0.024 \\
   2 & < 0.001 & < 0.001 & < 0.001 & < 0.001 & 0.003 & 0.021 & 0.001 & 0.004 \\
   3 & < 0.001 & < 0.001 & < 0.001 & < 0.001 & < 0.001 & 0.004 & < 0.001 & 0.002 \\
   4 & 0. & 0. & 0. & 0. & < 0.001 & 0.001 & < 0.001 & < 0.001 \\
   5 & 0. & 0. & 0. & 0. & < 0.001 & < 0.001 & < 0.001 & < 0.001 \\
   6 & 0. & 0. & 0. & 0. & < 0.001 & < 0.001 & 0. & 0. \\
   7 & 0. & 0. & 0. & 0. & - & - & - & - \\    
    \bottomrule
  \end{tabular}
  }  
\end{table} 

\begin{table}
  \caption{$\CauchyMinValidQBin$ on $\calibrationSplit$ for the standard multi-class classification experiments. The more challenging $\datasetFactcheck$ task has a commensurately higher $\CauchyMinValidQBin$.} 
  \label{tab:experiments-classification-min-valid-qbin-iterated} 
  \centering
  \resizebox{0.25\textwidth}{!}{%
  \begin{tabular}{ c c c c }
    \toprule
\multicolumn{2}{c}{$\datasetSentiment$} & \multicolumn{2}{c}{$\datasetFactcheck$} \\
     \cmidrule(r){1-2} \cmidrule(r){3-4}\\
    $\CauchyMAD$ & $\CauchyMinValidQBin$    & $\CauchyMAD$ &  $\CauchyMinValidQBin$  \\
    \midrule
      7.9e-05 & 1.004 & 0.100 &  2.447  \\
    \bottomrule
  \end{tabular}
  }  
\end{table} 

Table~\ref{tab:experiments-classification-combined} displays the results for the binary classification tasks. The results for $\datasetSentiment$ vs. those of the other datasets are indicative of the under-appreciated point in the existing calibration literature of the importance of comparisons over---at least modest---distribution-shifts. On in-distribution benchmark data with high accuracy models, the differences can be difficult to discern; after all, the class-wise accuracy of the model is itself $\ge \alpha'$. However, even in these otherwise straightforward binary classification settings, the existing classes of estimators all but fall apart in the presence of distribution shifts, which are common in practice with high-dimensional data, such as text. In this light, the existing classes of estimators are not demonstrably more effective than simply using an un-calibrated threshold on the output ($\estimatorSoftmax$). In contrast, the $\EstimatorPLower$ estimator achieves index-conditional calibration in all cases, correctly rejecting documents over which the estimates are unreliable, and admitting points for which the class- and prediction-conditional accuracies are $\ge \alpha'$.

Central to the unique behavior of the $\sdm$ estimator is that the epistemic uncertainty decreases as $\softqbin$ increases. Furthermore, $\hardqbin$ can be used as a mapping between $\calibrationSplit$ and a new, unseen test point, because the variation among comparable points also decreases as $\softqbin$ increases. Table~\ref{tab:experiments-classification-hard-qbin-iterated} shows this for the standard multi-class classification tasks with summary statistics over the $J=10$ iterations. The corresponding $\CauchyMinValidQBin$ used by the $\EstimatorPLower$ estimator (Eq.~\ref{eq:index-conditional-sdm-estimator}) appears in Table~\ref{tab:experiments-classification-min-valid-qbin-iterated}. Comparing these $\CauchyMinValidQBin$ values with Table~\ref{tab:experiments-classification-hard-qbin-iterated} makes it clear that the $\CauchyMinValidQBin$ values are effectively change points w.r.t. the uncertainty: Points below have high variation and points above have increasingly low variation to the point that $\CauchyMagnitudeIteratedLowerEstimate$ reaches 0, within numerical error. 

\textit{This behavior is remarkable for an estimator over high-dimensional inputs, because it demonstrates there are regions of the distribution that are low variation and high-probability that can be reliably detected.} Existing estimators marginalize over the distinctions in these regions, which can cause unexpected behavior at test time, as demonstrated in our empirical results.

\subsection{Results: Black-box LLM APIs}

\begin{table}
  \caption{Comparison of relevant estimators combined with $\modelGPTFourO$, $\alpha'=\colorbox{correctPredictionColor}{0.95}$. The \textsc{sdm} estimator, $\EstimatorPLower$, remains well-calibrated even over the much more challenging $\datasetMMLUPro4qa$ dataset. Importantly, $\EstimatorPLower$ is not vacuously conservative; the yield of admitted points is higher on $\datasetMMLU$ even when the verbalized uncertainty of $\modelGPTFourO$ is well-calibrated (see \underline{underline}). 
  }
  \label{tab:experiments-llm-api} 
  \centering
  \resizebox{0.65\textwidth}{!}{%
  \begin{tabular}{l l l c c }
    \toprule

Dataset  & Model & Estimator  & Acc. & $\frac{|\text{Admitted}|}{|\gD_{\rm{te}}|}$  \\
    \midrule
    $\datasetMMLU$ & $\modelGPTFourO$ & $\estimatorNoReject$  & \colorbox{wrongPredictionColor}{0.832} & 1.~~~~~  \\
    $\datasetMMLU$ & $\modelGPTFourO$  & $\estimatorLetterProb$  & $\colorbox{wrongPredictionColor}{0.921}$ & $0.74$    \\
    $\datasetMMLU$ & $\modelGPTFourO$ & $\estimatorVerbProb$  & $\colorbox{correctPredictionColor}{0.953}$ & $\underline{0.35}$   \\
    $\datasetMMLU$ & $\modelGPTFourOSDM$ & $\estimatorNoReject$   & \colorbox{wrongPredictionColor}{0.835} & 1.~~~~~   \\
    $\datasetMMLU$ & $\modelGPTFourOSDM$ & $\EstimatorPLower$  & $\colorbox{correctPredictionColor}{0.957}$ & $\underline{0.38}$   \\
    $\datasetMMLU$ & $\modelGPTFourOSDM$ & $\EstimatorPCentroid$  & $\colorbox{correctPredictionColor}{0.956}$ & $0.39$   \\
    $\datasetMMLU$ & $\modelGPTFourOSDM$ & $\EstimatorPUpper$  & $\colorbox{correctPredictionColor}{0.954}$ & $0.41$   \\
    \midrule
    $\datasetMMLUPro4qa$ & $\modelGPTFourO$ & $\estimatorNoReject$  & \colorbox{wrongPredictionColor}{0.648} & 1.~~~~~   \\
    $\datasetMMLUPro4qa$ & $\modelGPTFourO$  & $\estimatorLetterProb$  & $\colorbox{wrongPredictionColor}{0.870}$ & $0.51$    \\
    $\datasetMMLUPro4qa$ & $\modelGPTFourO$ & $\estimatorVerbProb$  & $\colorbox{wrongPredictionColor}{0.857}$ & $0.16$   \\
    $\datasetMMLUPro4qa$ & $\modelGPTFourOSDM$ & $\estimatorNoReject$   & \colorbox{wrongPredictionColor}{0.683} & 1.~~~~~   \\
    $\datasetMMLUPro4qa$ & $\modelGPTFourOSDM$ & $\EstimatorPLower$  & $\colorbox{correctPredictionColor}{0.958}$ & $0.22$   \\
    $\datasetMMLUPro4qa$ & $\modelGPTFourOSDM$ & $\EstimatorPCentroid$  & $\colorbox{correctPredictionColor}{0.957}$ & $0.23$  \\
    $\datasetMMLUPro4qa$ & $\modelGPTFourOSDM$ & $\EstimatorPUpper$  & $\colorbox{wrongPredictionColor}{0.942}$ & $0.24$   \\
    \bottomrule
  \end{tabular}
  }  
\end{table}

Table~\ref{tab:experiments-llm-api} contains the results of the estimators over $\modelGPTFourO$, the baseline accuracy (see $\estimatorNoReject$) of which is in-line with existing reported results for the zero-shot setting, and $\modelGPTFourOSDM$. Neither $\estimatorLetterProb$ nor $\estimatorVerbProb$ are reliable estimators across these datasets, even though the multiple-choice QA task is a common setting for LLM development and evaluation. Conceptually, both can be viewed as encoding the output $\Magnitude$, without explicitly controlling for the $\Similarity$ and $\Distance$, as with a $\estimatorSoftmax$ estimator in a standard classification setting. Their over-confidence on $\datasetMMLUPro4qa$ reflect this.

The results of $\EstimatorPLower$ on $\datasetMMLUPro4qa$ are indicative of the real-world use of the $\sdm$ estimator. $\modelGPTFourO$ has a dramatically lower overall accuracy on the $\datasetMMLUPro4qa$ questions, which would come as a surprise to an end-user who was expecting behavior similar to that over $\datasetMMLU$. In contrast, the $\EstimatorPLower$ estimator remains calibrated. For the rejected documents, the user would then know to take additional action. Alternatively, if part of an automated pipeline, additional test-time compute-based branching decisions (such as re-asking the model, or seeking outside information via retrieval) could be taken in the background before presenting a final result.

\paragraph{Data Quality Analysis.}

For \textsc{MMLU-Pro-4qa}, we examine the 5 questions in the Computer Science category that were in the $\EstimatorPLower$ index-conditional admitted set, but for which the predicted answers do not match the ground-truth annotations, $y \ne \hat{y}$. The top 4 questions sorted by $\EstimatorPLower$, all of which have $\EstimatorPLower \ge 0.99$, all clearly have annotation errors where the model predictions are correct and the ground-truth annotations are incorrect. We include the question id's in Table~\ref{tab:mmlu-pro-annotation-errors-cs}. This provides an exogenous evaluation of the method: The $\sdm$ estimator has successfully separated the aleatoric and epistemic uncertainty among the high-probability predictions.

\subsection{Results: Verified Generation}

\begin{table}
  \caption{Verified generation results, $\alpha'=\colorbox{correctPredictionColor}{0.95}$. Task datasets are identical to those in Table~\ref{tab:experiments-classification-combined}. Predictions are parsed from the JSON \textit{generated} by the model, with parsing errors counted as wrong predictions. \colorbox{correctPredictionColor}{N/A} indicates all predictions were rejected, which is preferred over falling under the expected accuracy. Verification via an $\sdm$ estimator is reliable regardless of fine-tuning the model, but fine-tuning with $\sdm$ ($\modelPhiThreeFiveInstructSDMNetwork$) can increase the task accuracy  (see \textbf{bold}) and the yield of admitted points (see \underline{underline}).} 
  \label{tab:experiments-verified-generation} 
  \centering
  \resizebox{0.65\textwidth}{!}{%
  \begin{tabular}{l l  l c c }
    \toprule
    Dataset   & Model & Estimator & \textsc{Acc.} & $\frac{|\text{Admitted}|}{|\testSplit|}$\\
    \midrule
    $\datasetSentiment$ & $\modelPhiThreeFiveInstructSDM$ & $\estimatorNoReject$ & \colorbox{wrongPredictionColor}{0.751} & 1.~~~~~ \\
    
    $\datasetSentiment$ & $\modelPhiThreeFiveInstructSDM$ & $\EstimatorPLower$ & \colorbox{correctPredictionColor}{0.997} & \underline{0.39} \\

    $\datasetSentiment$ & $\modelPhiThreeFiveInstructSDMNetwork$ & $\estimatorNoReject$  & \colorbox{wrongPredictionColor}{\textbf{0.876}} & 1.~~~~~ \\
    
    $\datasetSentiment$ & $\modelPhiThreeFiveInstructSDMNetwork$ & $\EstimatorPLower$ & \colorbox{correctPredictionColor}{0.996} & \underline{0.42} \\

  \midrule 
  
  $\datasetSentimentOOD$ & $\modelPhiThreeFiveInstructSDM$ & $\estimatorNoReject$  & \colorbox{wrongPredictionColor}{0.815} & 1.~~~~~ \\
  
  $\datasetSentimentOOD$ & $\modelPhiThreeFiveInstructSDM$ & $\EstimatorPLower$ & \colorbox{correctPredictionColor}{1.~~~~~~} & <0.01~~~ \\

    $\datasetSentimentOOD$ & $\modelPhiThreeFiveInstructSDMNetwork$ & $\estimatorNoReject$ & \colorbox{wrongPredictionColor}{\textbf{0.896}} & 1.~~~~~ \\
  
  $\datasetSentimentOOD$ & $\modelPhiThreeFiveInstructSDMNetwork$ & $\EstimatorPLower$ & \colorbox{correctPredictionColor}{1.~~~~~~} & <0.01~~~ \\

  \midrule 

 $\datasetFactcheck$ & $\modelPhiThreeFiveInstructSDM$ & $\estimatorNoReject$ & \colorbox{wrongPredictionColor}{0.706} & 1.~~~~~ \\
 
 $\datasetFactcheck$ & $\modelPhiThreeFiveInstructSDM$ & $\EstimatorPLower$ & \colorbox{correctPredictionColor}{0.973} & 0.15  \\

  $\datasetFactcheck$ & $\modelPhiThreeFiveInstructSDMNetwork$ & $\estimatorNoReject$ & \colorbox{wrongPredictionColor}{\textbf{0.743}} & 1.~~~~~ \\
 
 $\datasetFactcheck$ & $\modelPhiThreeFiveInstructSDMNetwork$ & $\EstimatorPLower$ & \colorbox{correctPredictionColor}{0.973} & 0.15 \\

    \bottomrule
  \end{tabular}
  }  
\end{table}

The results for the $\sdm$ network indicate effective verification of instruction following (Table~\ref{tab:experiments-verified-generation}). Our small-scale experiment confirms that the $\sdmVerificationLayer$ reliably yields a calibrated estimator regardless of fine-tuning, but the fine-tuning process improves overall task accuracy. That is, the results confirm that Alg.~\ref{alg:sdm-network-training}, which chose epoch 3 of 5 as the final model, is a viable fine-tuning loss and process. Importantly in this context, the cardinality of the set of admitted points is non-decreasing relative to before fine-tuning, despite updating 100 million parameters on a small training set. Leveraging the behavior of the $\sdm$ estimator, the $\sdm$ network is, in this way, the first statistically principled and robust approach to construct an LLM with an intrinsic ability to verify its own instruction-following and generated output.

\section{Conclusion}
There has been renewed interest in deep learning as a focus of research for language modeling over the last decade, and a growing number of efforts to scale data and model compute for various applications. However, brittleness to distribution shifts, lack of reliable uncertainty quantification, and opaque predictions with respect to the training data have precluded---or otherwise diminished the potential of---the use of neural network language models in most real-world settings. In this work, we have addressed these foundational limitations by introducing $\sdm$ activation functions, $\sdm$ calibration, and $\sdm$ networks.




\newpage

\bibliography{sdm}
\bibliographystyle{ml_format}

\newpage
\appendix
\section{Appendix}
We provide additional experimental details and results for the black-box LLM API experiments in \S~\ref{Appendix:LLM-APIs} and the verified generation experiments in \S~\ref{Appendix:verified-generation}. Additional training details are included in \S~\ref{appendix:additional-training-details}.

Code to replicate our results is available at the URL provided in the main text. For the reader, we provide a few key highlights here. We include an implementation of the $\sdm$ activation function in \S~\ref{appendix-sdm-activation-implementation}. We provide our conventions for calculating empirical CDFs in \S~\ref{appendix:empirical-cdf-implementation}, and we provide code scaffolding for an example implementation of an $\sdm$ network training loop in \S~\ref{appendix:neg+pos-normalization-and-reg}.

\subsection{Black-box LLM APIs}
\label{Appendix:LLM-APIs}
The results of the data quality analysis are included in Table~\ref{tab:mmlu-pro-annotation-errors-cs}. Following best practices, to avoid contaminating the test set since research articles are commonly used for LLM training, we only include the question id's and not the question and answer text, which can readily be retrieved from the Huggingface datasets database.

We include the prompts used for the experiments in the code repo. The prompt is a variation on the theme of that used in OpenAI's Simple Evals repo\footnote{\url{https://github.com/openai/simple-evals/blob/main/common.py}}, with the addition of using structured outputs against the JSON Schema in Listing~\ref{lst:llm-json-schema}. The particular prompt and structuring of the JSON (and parsing of the JSON, described below) are not defining aspects of the approach and are not necessarily the optimal templates. We use a direct, zero-shot approach to examine the more challenging setting---arguably closer to real-world usage---than providing examples or systematically hill-climbing on prompts.

The embedding for input to the \textsc{sdm} activation layer is constructed by parsing the JSON schema mapped back to the top-1 probabilities of the output tokens. For each key, we average the log-probabilities in probability space of the tokens of the corresponding value. For example, for the key \texttt{"short\_explanation\_for\_answer\_confidence"}, we parse the output to isolate the tokens corresponding to the value, and take the average of the exponentiated log probabilities of the tokens. Given the 3 keys in the JSON schema, this results in 3 floating-point values. The verbalized uncertainty key \texttt{"confidence\_in\_answer\_letter"} has a value of type \texttt{number}, but the output itself corresponds to a sequence of discrete tokens (e.g., ``\texttt{0}'', ``\texttt{.}'', ``\texttt{9}''), so this parsing process is the same as that for the values of type \texttt{string}. Finally, we construct a soft one-hot vector of length 4 where the non-zero index (if any) of the predicted letter is set to the floating-point value of the verbalized uncertainty (i.e., the value for the key \texttt{"confidence\_in\_answer\_letter"}). The input embedding is then the concatenation of these 7 values. Full refusals from the LLM's API, which are rare but can occur on some of the social science and humanities questions, are assigned vectors of 0's as embeddings for $\trainSplit$ and $\calibrationSplit$ instances, and treated as wrong predictions in the test evaluations.

The estimator $\estimatorLetterProb$ corresponds to the index of this embedding derived from the value of the key \texttt{"answer\_letter"}. Often this is the probability of the single token (i.e., ``\texttt{A}'', ``\texttt{B}'', ``\texttt{C}'', ``\texttt{D}''), but occasionally will be the average over additional tokens (e.g., ``\texttt{\$}''). The estimator $\estimatorVerbProb$ corresponds to the floating-point value of the verbalized uncertainty.

In our experiments, we aim for a controlled comparison with $\estimatorLetterProb$ and $\estimatorVerbProb$; as such, the \textsc{sdm} activation layer is only given access to the 7 values above. In particular, we do not provide access to additional signal derived from composition with another model. In applications where the uncertainty is over multiple tasks (i.e., not just question answering of this particular format), to avoid a marginalization over tasks, we recommend either encoding the distinction across tasks in the JSON schema, or simply concatenating the LLM output with the hidden states of another large model. The latter is typically readily achievable by running another model alongside the black-box LLM's API.

We train the \textsc{sdm} activation layer as a 4-class classification task, which is an effective but potentially sample-inefficient encoding, at least when assuming the absence of artifacts correlated with answer letters. An alternative would be to re-encode the task as binary classification, either as a leave-one-out classification or as binary verification (as in \S~\ref{ref:experiments-universal-verification}). Since the choice of encoding, as with the structure of the prompt and JSON Schema, is orthogonal to the evaluation of the uncertainty estimates---other than with respect to effective sample sizes---we keep these aspects straightforward in this set of experiments to avoid complicating the presentation.

Given the results in the main text, a next step would be to use this behavior to build a \textit{re-ask} pipeline. That is, predictions with low probability can be automatically routed to re-prompt the LLM conditional on the previous response, a potentially effective means of building test-time compute systems over otherwise black-box models. Such pipelines are not feasible without robust estimates of predictive uncertainty, but become conceptually straightforward---and straightforward to implement---given the behavior of $\sdm$ estimators. We leave such additional applied examples for future work to systematically analyze.

\begin{table}
\caption{$\datasetMMLUPro4qa$, Computer Science category. Predictions that met the index-conditional threshold but were marked incorrect according to the ground-truth labels. Examination of the data reveals the model is \colorbox{correctPredictionColor}{correct} and the ground-truth annotations are \colorbox{wrongPredictionColor}{incorrect}. The digit significance of $\EstimatorP$ is not necessarily significant (and when shown to users, would typically be rounded, with a top ceiling to avoid 1.0), but provided for reference. $\nYhat$ is the effective sample size for the predicted class. The final question is arguably ambiguous.} 
  \label{tab:mmlu-pro-annotation-errors-cs} %
  \centering
  \resizebox{\textwidth}{!}{%
  \begin{tabular}{l c c l l l l }
    \toprule

Question ID   & y & $\hat{y}$  & $\EstimatorPLower$ & $\EstimatorPCentroid$ & $\EstimatorPUpper$ & $\nYhat$   \\
    \midrule
    $10750$ & \colorbox{wrongPredictionColor}{A} & \colorbox{correctPredictionColor}{D} & 0.9999999029119694 & 0.999999946869715 & 0.999999963752475 & 11563  \\
    $10682$ & \colorbox{wrongPredictionColor}{D}  & \colorbox{correctPredictionColor}{C} & 0.9999995410050875 & 0.9999997504521737 & 0.9999998413548795 & 11774  \\
    $10458$ & \colorbox{wrongPredictionColor}{D} & \colorbox{correctPredictionColor}{A} & 0.9997548501324156 & 0.9998610348657851 & 0.9999170919091862 & 9129 \\
    $10533$ & \colorbox{wrongPredictionColor}{B} & \colorbox{correctPredictionColor}{C} & 0.9897059289074643 & 0.9936086673736274 & 0.9957749405311342 & 6891 \\
    $10479$ & D & B & 0.967751071803557 & 0.9791966686070331 & 0.9862756083406558 & 7684 \\
    \bottomrule
  \end{tabular}
  }  
\end{table}
 
\begin{ListingWithCaption}
\begin{lstlisting}[caption={JSON Schema for $\modelGPTFourO$ Structured Outputs.}, label={lst:llm-json-schema}]
{
    "properties": {
        "answer_letter": {
            "title": "Answer Letter",
            "type": "string"
        },
        "confidence_in_answer_letter": {
            "title": "Confidence In Answer Letter",
            "type": "number"
        },
        "short_explanation_for_answer_confidence": {
            "title": "Short Explanation For Answer Confidence",
            "type": "string"
        }
    },
    "required": [
        "answer_letter",
        "confidence_in_answer_letter",
        "short_explanation_for_answer_confidence"
    ],
    "title": "MultipleChoiceQuestionResponse",
    "type": "object"
}
\end{lstlisting}
\end{ListingWithCaption}

\subsection{Verified Generation}
\label{Appendix:verified-generation}

For reference, Table~\ref{tab:experiments-verification} provides the effectiveness over the force-decoded datasets. The support set of the $\sdmVerificationLayer$ is constructed from the force-decoded training and calibration data, so this table reflects the held-out classification ability over the verification data, which includes constructed negatives for $\yVerification=0$, as described in the main text and illustrated in Table~\ref{tab:experiments-verified-generation-data-format}. Listing~\ref{lst:verified-generation-inputs} includes the system message and prompts used for the experiments. 

\begin{table}
  \caption{Verification results on the \textit{force-decoded test sets} for reference, $\alpha'=\colorbox{correctPredictionColor}{0.95}$. See Table~\ref{tab:experiments-verified-generation} for generation results for the underlying tasks, which reflect real test-time usage. \colorbox{correctPredictionColor}{N/A} indicates all predictions were rejected, which is preferred over falling under the expected accuracy. 
  $n=|\text{Admitted}|$, the count of non-rejected documents. Additional resolution added to $\frac{n}{|\testSplit|}$ columns for $\datasetSentimentOODVerification$ for reference, but the number of admitted points is effectively 0.} %
  \label{tab:experiments-verification} 
  \centering
  \resizebox{1.0\textwidth}{!}{%
  \begin{tabular}{l l  l c c c c c c c c c l }
    \toprule

    & & &  \multicolumn{4}{c}{Class-conditional} & \multicolumn{4}{c}{Prediction-conditional}  & \multicolumn{2}{c}{Marginal} \\
    & & &  \multicolumn{2}{c}{$y=0$} & \multicolumn{2}{c}{$y=1$} & \multicolumn{2}{c}{$\hat{y}=0$} & \multicolumn{2}{c}{$\hat{y}=1$} & \multicolumn{2}{c}{$y\in \{0,1\}$} \\
    \cmidrule(r){4-5} \cmidrule(r){6-7} \cmidrule(r){8-9} \cmidrule(r){10-11} \cmidrule(r){12-13} \\
    Dataset   & Model & Estimator & \textsc{Acc.}& $\frac{n}{|\testSplit|}$ & \textsc{Acc.} & $\frac{n}{|\testSplit|}$ & \textsc{Acc.} & $\frac{n}{|\testSplit|}$ & \textsc{Acc.} & $\frac{n}{|\testSplit|}$ & \textsc{Acc.} & $\frac{n}{|\testSplit|}$\\
    \midrule
    $\datasetSentimentVerification$ & $\modelPhiThreeFiveInstructSDM$ & $\estimatorNoReject$ & \colorbox{correctPredictionColor}{0.959} & 0.51 & \colorbox{wrongPredictionColor}{0.891} & 0.49 & \colorbox{wrongPredictionColor}{0.901} & 0.54 & \colorbox{correctPredictionColor}{0.954} & 0.46 & \colorbox{wrongPredictionColor}{0.925} & 1. \\
    
    $\datasetSentimentVerification$ & $\modelPhiThreeFiveInstructSDM$ & $\EstimatorPLower$ & \colorbox{correctPredictionColor}{0.996} & 0.17 & \colorbox{correctPredictionColor}{0.997} & 0.21 & \colorbox{correctPredictionColor}{0.996} & 0.17 & \colorbox{correctPredictionColor}{0.997} & 0.21 & \colorbox{correctPredictionColor}{0.997} & 0.38 \\
    
    $\datasetSentimentVerification$ & $\modelPhiThreeFiveInstructSDM$ & $\EstimatorPCentroid$ & \colorbox{correctPredictionColor}{0.996} & 0.18 & \colorbox{correctPredictionColor}{0.997} & 0.22 & \colorbox{correctPredictionColor}{0.996} & 0.18 & \colorbox{correctPredictionColor}{0.997} & 0.22 & \colorbox{correctPredictionColor}{0.997} & 0.40 \\
    
    $\datasetSentimentVerification$ & $\modelPhiThreeFiveInstructSDM$ & $\EstimatorPUpper$ & \colorbox{correctPredictionColor}{0.997} & 0.19 & \colorbox{correctPredictionColor}{0.997} & 0.23 & \colorbox{correctPredictionColor}{0.997} & 0.19 & \colorbox{correctPredictionColor}{0.997} & 0.23 & \colorbox{correctPredictionColor}{0.997} & 0.42 \\
  \midrule 
  
  $\datasetSentimentOODVerification$ & $\modelPhiThreeFiveInstructSDM$ & $\estimatorNoReject$ & \colorbox{correctPredictionColor}{0.978} & 0.51 & \colorbox{wrongPredictionColor}{0.639} &0.49 & \colorbox{wrongPredictionColor}{0.738} & 0.68 & \colorbox{correctPredictionColor}{0.966} & 0.32 & \colorbox{wrongPredictionColor}{0.812} & 1. \\
  
  $\datasetSentimentOODVerification$ & $\modelPhiThreeFiveInstructSDM$ & $\EstimatorPLower$ & \colorbox{correctPredictionColor}{1.~~~~~~} & 0.002 & \colorbox{correctPredictionColor}{1.~~~~~~} & 0.0002 & \colorbox{correctPredictionColor}{1.~~~~~~} & 0.002 & \colorbox{correctPredictionColor}{1.~~~~~~} & 0.0002 & \colorbox{correctPredictionColor}{1.~~~~~~} & 0.002 \\
  $\datasetSentimentOODVerification$ & $\modelPhiThreeFiveInstructSDM$ & $\EstimatorPCentroid$ & \colorbox{correctPredictionColor}{1.~~~~~~} & 0.003 & \colorbox{correctPredictionColor}{1.~~~~~~} & 0.0002 & \colorbox{correctPredictionColor}{1.~~~~~~} & 0.003 & \colorbox{correctPredictionColor}{1.~~~~~~} & 0.0002 & \colorbox{correctPredictionColor}{1.~~~~~~} & 0.003 \\
  $\datasetSentimentOODVerification$ & $\modelPhiThreeFiveInstructSDM$ & $\EstimatorPUpper$ & \colorbox{correctPredictionColor}{1.~~~~~~} & 0.003 & \colorbox{correctPredictionColor}{1.~~~~~~} & 0.0004 & \colorbox{correctPredictionColor}{1.~~~~~~} & 0.003 & \colorbox{correctPredictionColor}{1.~~~~~~} & 0.0004 & \colorbox{correctPredictionColor}{1.~~~~~~} & 0.004 \\
  \midrule 

 $\datasetFactcheckVerification$ & $\modelPhiThreeFiveInstructSDM$ & $\estimatorNoReject$ & \colorbox{wrongPredictionColor}{0.656} & 0.50 & \colorbox{wrongPredictionColor}{0.732} & 0.50 & \colorbox{wrongPredictionColor}{0.708} & 0.46 & \colorbox{wrongPredictionColor}{0.682} & 0.54 & \colorbox{wrongPredictionColor}{0.694} & 1. \\
 
 $\datasetFactcheckVerification$ & $\modelPhiThreeFiveInstructSDM$ & $\EstimatorPLower$ & \colorbox{correctPredictionColor}{~N/A~} & 0.~~~~ & \colorbox{correctPredictionColor}{1.~~~~~~} & 0.07 & \colorbox{correctPredictionColor}{~N/A~} & 0.~~~~ & \colorbox{correctPredictionColor}{1.~~~~~~} & 0.07 & \colorbox{correctPredictionColor}{1.~~~~~~} & 0.07 \\
 
 $\datasetFactcheckVerification$ & $\modelPhiThreeFiveInstructSDM$ & $\EstimatorPCentroid$ & \colorbox{correctPredictionColor}{~N/A~} & 0.~~~~ & \colorbox{correctPredictionColor}{1.~~~~~~} & 0.08 & \colorbox{correctPredictionColor}{~N/A~} & 0.~~~~ & \colorbox{correctPredictionColor}{1.~~~~~~} & 0.08 & \colorbox{correctPredictionColor}{1.~~~~~~} & 0.08 \\
  
 $\datasetFactcheckVerification$ & $\modelPhiThreeFiveInstructSDM$ & $\EstimatorPUpper$ & \colorbox{correctPredictionColor}{~N/A~} & 0.~~~~ & \colorbox{correctPredictionColor}{1.~~~~~~} & 0.08 & \colorbox{correctPredictionColor}{~N/A~} & 0.~~~~ & \colorbox{correctPredictionColor}{1.~~~~~~} & 0.08 & \colorbox{correctPredictionColor}{1.~~~~~~} & 0.08 \\
   
    \bottomrule
  \end{tabular}
  }  
\end{table}

\begin{table}
  \caption{JSON structure for the verified generation experiments, with $\gM^{\rm{ref}}= \modelPhiThreeFiveInstruct$. $\yVerification=1$ corresponds to the standard classification tasks, where, e.g., $\yTask=0$ corresponds to a negative review for the sentiment task, and $\yTask=1$ corresponds to a factually correct statement for the factcheck task. $\yVerification=0$ flips the parity, and is used for constructing negatives for training, and the contrastive basis for rejection at test-time. Recall that the LLM takes as input a system prompt, user prompt, and the document (see Listing~\ref{lst:verified-generation-inputs}). At test time, we seek to generate the correct JSON output (i.e., that corresponding to the correct $\yTask$ label), for instances with $\yVerificationPrediction=1$ predicted by the $\sdmVerificationLayer$ layer.} 
  \label{tab:experiments-verified-generation-data-format} 
  \centering
  \resizebox{1.0\textwidth}{!}{%
  \begin{tabular}{l l l}
     \toprule
     Datasets   & Labels & JSON output \\
     \midrule
  
\rowcolor{lightestgray} $\datasetSentiment$, $\datasetSentimentVerification$ & & \\
\rowcolor{lightestgray} $\datasetSentimentOOD$, $\datasetSentimentOODVerification$ & & \\
 &
$\yTask=0 , \yVerification=1$ &
\ttfamily{
\{"sentiment": "negative"\}
} \\

 & 
$\yTask=1 , \yVerification=1$ &
\ttfamily{
\{"sentiment": "positive"\}
} \\

 & 
$\yTask=0 , \yVerification=0$ &
\ttfamily{
\{"sentiment": "positive"\}
} \\

 & 
$\yTask=1 , \yVerification=0$ &
\ttfamily{
\{"sentiment": "negative"\}
} \\

\rowcolor{lightestgray} $\datasetFactcheck$, $\datasetFactcheckVerification$ & & \\
 &
$\yTask=0 , \yVerification=1$ &
\ttfamily{
\{"correctness": false\}
} \\

 &
$\yTask=1 , \yVerification=1$ &
\ttfamily{
\{"correctness": true\}
} \\

 &
$\yTask=0 , \yVerification=0$ &
\ttfamily{
\{"correctness": true\}
} \\

 &
$\yTask=1 , \yVerification=0$ &
\ttfamily{
\{"correctness": false\}
} \\

    \bottomrule
  \end{tabular}
  }  
\end{table}

\begin{ListingWithCaption}
\begin{lstlisting}[caption={System and user messages for the sentiment and factcheck datasets of the verified generation experiments, with $\gM^{\rm{ref}}= \modelPhiThreeFiveInstruct$. The document text replaces TEXT for each instance.}, label={lst:verified-generation-inputs}]
<|system|>
You are a helpful AI assistant.<|end|>
<|user|>
Classify the sentiment of the following movie review. Respond using the following JSON: {"sentiment": str}. REVIEW: TEXT<|end|>
<|assistant|>

<|system|>
You are a helpful AI assistant.<|end|>
<|user|>
Check the following document for hallucinations and/or factual inaccuracies. Respond using the following JSON: {"correctness": bool}. DOCUMENT: TEXT<|end|>
<|assistant|>
\end{lstlisting}
\end{ListingWithCaption}

\subsection{Additional Training Details}\label{appendix:additional-training-details}

\paragraph{Compute.} The black-box LLM experiments require API calls, as detailed in the main text, but all other results can be reproduced locally on a single 2023 Mac Studio with an M2 Ultra chip with 128 GB of unified memory. These experiments are designed to fully assess the methods while still being replicable with consumer hardware.

\paragraph{Hyper-parameters.}
In the code repo, we include scripts for replicating our results. For all cases, we train the rescaling transform (Alg.~\ref{alg:train-rescaler}) for up to 1000 epochs, with early stopping if the loss exceeds the min observed loss for 10 consecutive epochs. In all experiments, $M=1000$ and we use a mini-batch size of 50. We mean center the input to $g$, the 1-D CNN of the $\sdm$ activation layer, via the mean and standard deviation over $\trainSplit$. We train $\modelGPTFourOSDM$ for $J=10$ iterations of 5 epochs, and the $\sdm$ models of $\datasetSentiment$ and $\datasetFactcheck$, as well as the $\sdmVerificationLayer$ of the $\sdm$ network, for $J=10$ iterations of 50 epochs. The standard exemplar adaptors of the $\datasetSentiment$ and $\datasetFactcheck$ classification experiments are trained with cross-entropy losses for 50 epochs. We use the Adam optimizer \citep{Kingma-2017-Adam-Optimizer} with a learning rate of $1 \times 10^{-4}$ for training the rescaling transform (Alg.~\ref{alg:train-rescaler}) and $1 \times 10^{-5}$ for all other cases.

\subsection{Example Implementation of the SDM Activation Function}\label{appendix-sdm-activation-implementation}
We include an implementation of the $\sdm$ activation function using PyTorch \citep{PaszkeEtAl-2019-PyTorch}, version 2.3.0, in Listing~\ref{lst:sdm-activation}.

\begin{ListingWithCaption}
\begin{lstlisting}[caption={Implementation of the $\sdm$ activation function in PyTorch, version 2.3.0.}, label={lst:sdm-activation}]
def sdm_activation_function(batch_input, q, distance_quantile_per_class=None, log=False):
    """
    sdm activation function
    Parameters
    ----------
    batch_input
        torch.tensor
            shape == [batch size, number of classes]
    q
        torch.tensor
            shape == [batch size, 1], with each value in [0, max q]
    distance_quantile_per_class
        torch.tensor, or None
            If not None, shape == [batch size, number of classes], with each quantile in [0,1]. As a final layer
            activation function, with batch_input $\in \reals$, it is assumed that the quantiles are the same
            across classes, for a given instance. This ensures the argmax does not change relative to
            torch.argmax(batch_input, dim=1).
    log
        log with change of base, for training
    Notes:
        For context, with e.g. batch size = 1, the standard softmax is obtained by using q=torch.tensor([[torch.e-2]])
        and (distance_quantile_per_class=None or distance_quantile_per_class=torch.ones(1, number of classes) ).
    Returns
    -------
    [batch size, number of classes]
    """
    assert len(batch_input.shape) == 2
    assert batch_input.shape[0] == q.shape[0]
    assert q.shape[1] == 1
    if distance_quantile_per_class is not None:
        assert batch_input.shape == distance_quantile_per_class.shape
    q_rescale_offset = 2
    q_factor = q_rescale_offset + q
    batch_input = batch_input - torch.amax(batch_input, dim=1, keepdim=True)  # for numerical stability
    if distance_quantile_per_class is not None:
        rescaled_distribution = q_factor ** (batch_input * distance_quantile_per_class)
    else:
        rescaled_distribution = q_factor ** batch_input
    if log:  # log_base{q}
        kEPS = torch.finfo(torch.float32).eps  # adjust as applicable for platform
        rescaled_distribution = torch.log(rescaled_distribution + kEPS) - torch.log(
            torch.sum(rescaled_distribution, dim=1) + kEPS).unsqueeze(1)
        return rescaled_distribution / torch.log(q_factor)
    else:
        return rescaled_distribution / torch.sum(rescaled_distribution, dim=1).unsqueeze(1)
\end{lstlisting}
\end{ListingWithCaption}

\raggedbottom
\subsection{Empirical CDF Function}\label{appendix:empirical-cdf-implementation}

\begin{ListingWithCaption}
\begin{lstlisting}[caption={An implementation of the empirical CDF conventions used in this work, using NumPy, version 1.26.4. See the text for a further discussion.}, label={lst:empirical-cdfs}]
    def getCDFIndex(trueClass_To_CDF, val, prediction, reverse=False, val_in_0to1=False):
    	# trueClass_To_CDF is a dictionary with a key for each class, the values of which are sorted ascending lists of numbers, since np.searchsorted assumes an ascending sort of its initial argument.
        if prediction not in trueClass_To_CDF or len(trueClass_To_CDF[prediction]) == 0:
            return 0.0
        if val_in_0to1 and len(trueClass_To_CDF[prediction]) > 0 and val >= trueClass_To_CDF[prediction][-1]:  # saturation guard
            assert not reverse
            return 1.0
        index = np.searchsorted(trueClass_To_CDF[prediction], val, side="left")  # will be 0 for len() == 0
        if reverse:  # use for distances
            return 1 - index / len(trueClass_To_CDF[prediction])
        else:
            return index / len(trueClass_To_CDF[prediction])
\end{lstlisting}
\end{ListingWithCaption}

The conventions for implementing the empirical CDF functions follow in the expected ways, but we briefly highlight the key considerations below, as they can impact the behavior of the estimators. An implementation in NumPy \citep{harrisEtAl-2020-Numpy}, version 1.26.4, appears in Listing~\ref{lst:empirical-cdfs}.

\begin{enumerate}
\item The distance quantiles should be exclusionary at the boundaries. When $\dNearest=0$, the $1-{\rm{eCDF}}^{\cdot}_{\rm{ca}}( \dNearest ) $ quantile should be 1, and when $\dNearest$ is greater than the maximum observed distance (across $\calibrationSplit$ for $\vx \in \testSplit$ and $\vx \in \calibrationSplit$, and across $\trainSplit$ for $\vx \in \trainSplit$, the latter case only occurring during training), the $1-{\rm{eCDF}}^{\cdot}_{\rm{ca}}( \dNearest ) $ quantile should be 0.
\item For the quantiles over an $\sdm$ activation, as needed for calibration, saturated values at the high-end should be assigned a quantile of 1. In the example code, this is achieved by setting the argument \texttt{val\_in\_0to1=True}.
\end{enumerate}

\pagebreak

\subsection{Example Implementation of the Negative+Positive Vocabulary Normalization and $L^2$ Regularization Term}\label{appendix:neg+pos-normalization-and-reg}
The positive+negative vocabulary normalization and regularization loss (Eq.~\ref{eq:sdm-network-next-token-training-loss}) are conceptually parsimonious and straightforward to implement. Code scaffolding for an example implementation of an $\sdm$ network training loop appears in Listing~\ref{lst:negative-positive-l2-epoch}. For computational expediency, here (as in the experiments in the main text), the $\q$ values and distance quantiles are calculated after each epoch, although in principle, they can be calculated with updated network values as an epoch progresses.

\begin{ListingWithCaption}
\begin{lstlisting}[caption={Code scaffolding in PyTorch, version 2.3.0, for a basic training loop of an $\sdm$ network with the Negative+Positive Vocabulary Normalization and $L^2$ regularization term, where the $\q$ values and distance quantiles are updated after each epoch.}, label={lst:negative-positive-l2-epoch}]
pdist = nn.PairwiseDistance(p=2)
criterion = nn.NLLLoss()
for e in range(total_epochs):
    total_mini_batches = len(range(0, train_size, batch_size))
    beta = min_beta 
    beta_step = (max_beta-min_beta) / total_mini_batches
    for i in range(0, train_size, batch_size):
        optimizer.zero_grad()
        model.train()
        batch_genai_y = # the next-token labels with applicable index+|V| offsets
        # the sdm activations for the negative+positive joint distribution and the concatenation of the reference
        # distribution with itself use the same q and distance quantiles for the corresponding instances:
        batch_f_genai = # log_base{q} sdm activation(negative+positive linear layers output), where + is pseudo-code for concatenation
        batch_f_original = # log_base{q} sdm activation(reference distribution+reference distribution linear layers output)
        with torch.no_grad():
            top_events_k = 1
            top_k_sort_by_largest = True
            # "negative" refers to indexes in the first half of the concatenated distributions, [0, |V|); "positive" to the second half [|V|, |V|*2):
            neg_original_max_half_distribution_i = torch.topk(batch_f_original[:, 0:model.gen_ai_vocab],
                                                              top_events_k, dim=1, largest=top_k_sort_by_largest)[1]
            pos_original_max_half_distribution_i = torch.topk(batch_f_original[:, -model.gen_ai_vocab:],
                                                              top_events_k, dim=1, largest=top_k_sort_by_largest)[1] + model.gen_ai_vocab  # note the offset
            negative_max_half_distribution_i = torch.topk(batch_f_genai[:, 0:model.gen_ai_vocab],
                                                          top_events_k, dim=1, largest=top_k_sort_by_largest)[1]
            positive_max_half_distribution_i = torch.topk(batch_f_genai[:, -model.gen_ai_vocab:],
                                                          top_events_k, dim=1, largest=top_k_sort_by_largest)[1] + model.gen_ai_vocab  # note the offset
            distribution_mass_mask = (
                    torch.ones_like(batch_f_genai).scatter_(1, neg_original_max_half_distribution_i, 0.0) *
                    torch.ones_like(batch_f_genai).scatter_(1, pos_original_max_half_distribution_i, 0.0) *
                    torch.ones_like(batch_f_genai).scatter_(1, negative_max_half_distribution_i, 0.0) *
                    torch.ones_like(batch_f_genai).scatter_(1, positive_max_half_distribution_i, 0.0) *
                    torch.ones_like(batch_f_genai).scatter_(1, batch_genai_y.unsqueeze(1), 0.0)
            ).to(batch_f_genai.device)
        regularization_term = pdist(
            distribution_mass_mask * batch_f_original,
            distribution_mass_mask * batch_f_genai).mean()
        llm_loss = criterion(batch_f_genai, batch_genai_y)
        with torch.no_grad():  # rescaling factor for the regularization term
            regularization_scale_term = (torch.log(llm_loss + model.kEPS) /
                                         (torch.log(regularization_term + model.kEPS) + model.kEPS)
                                         ).item()
        loss = llm_loss + beta * torch.sqrt(
            torch.clamp(regularization_term, min=1.0) ** min(max(regularization_scale_term, 0.0), 1.0))
        loss.backward()
        optimizer.step()
        beta += beta_step
    # Before the next epoch, for each training instance, update q and distance quantiles using the sdm activation layer trained for verification.
\end{lstlisting}
\end{ListingWithCaption}

\end{document}